\documentclass[11pt]{article}

\usepackage{xcolor,colortbl}
\usepackage[final]{acl}

\usepackage{times}
\usepackage{latexsym}

\usepackage[T1]{fontenc}

\usepackage[utf8]{inputenc}
\usepackage{scalerel}
\def\thinkingface{\scalerel*{\includegraphics{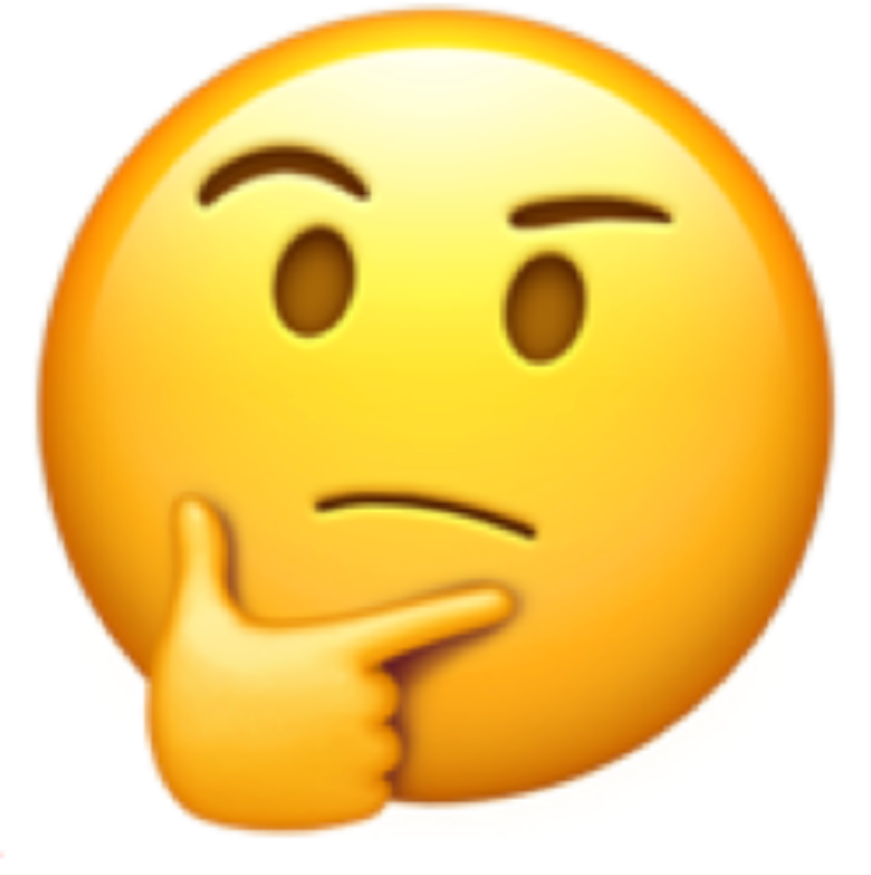}}{\strut}~}
\def\biceps{\scalerel*{\includegraphics{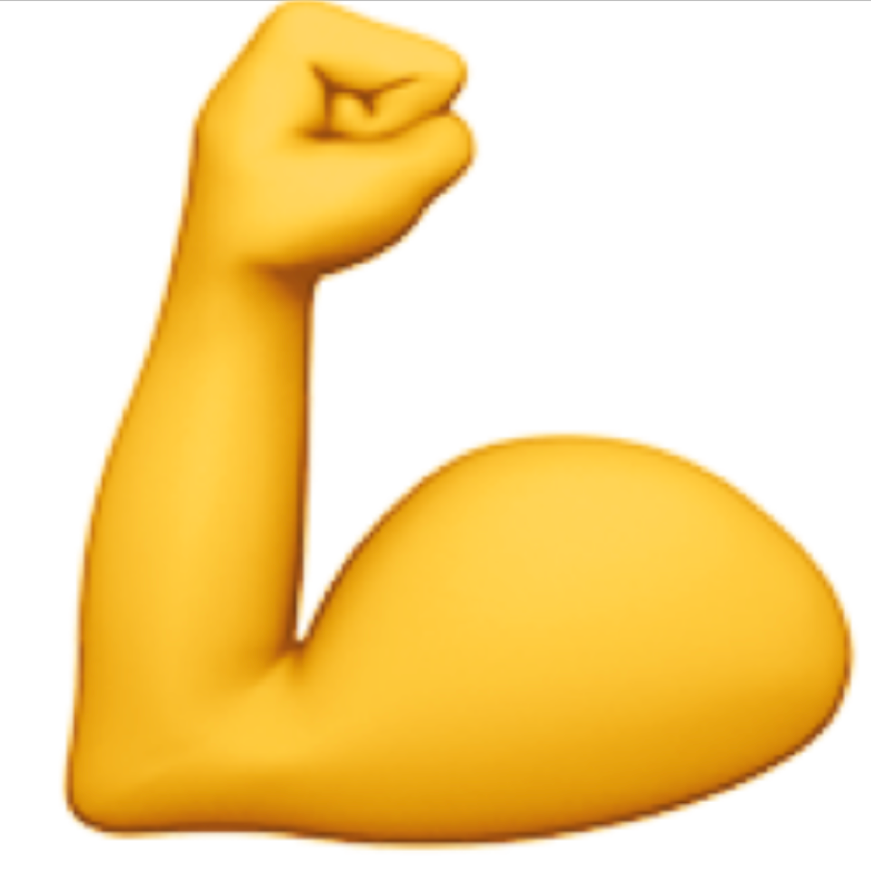}}{\strut}~}
\def\sparkles{\scalerel*{\includegraphics{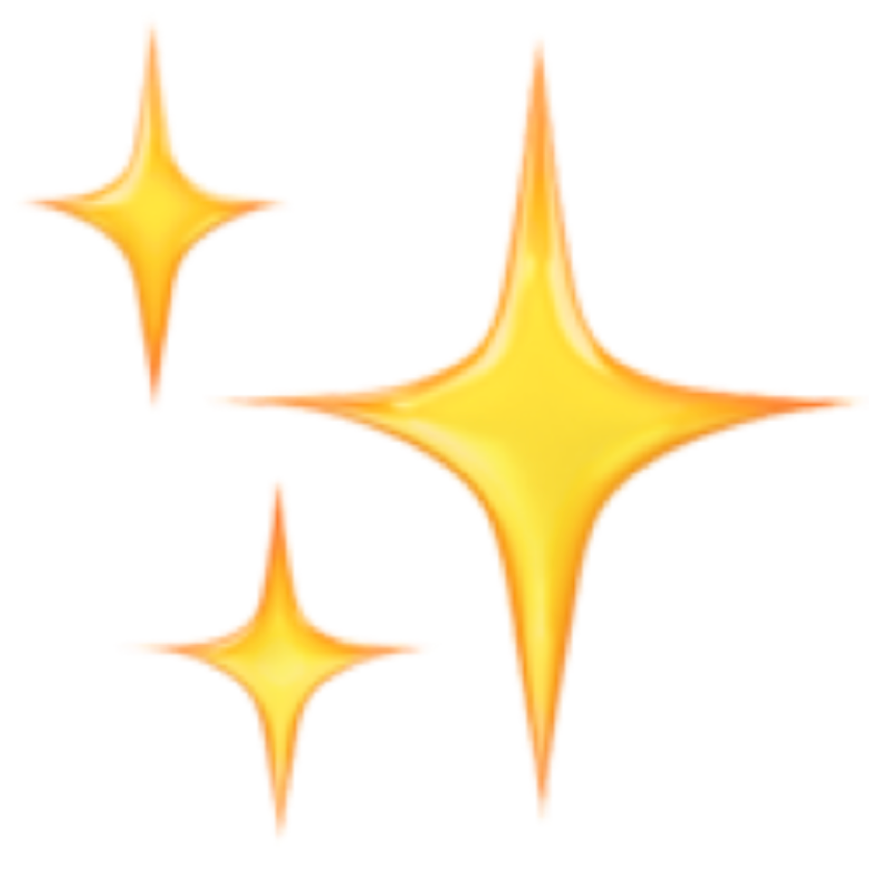}}{\strut}~}
\def\purpleheart{\scalerel*{\includegraphics{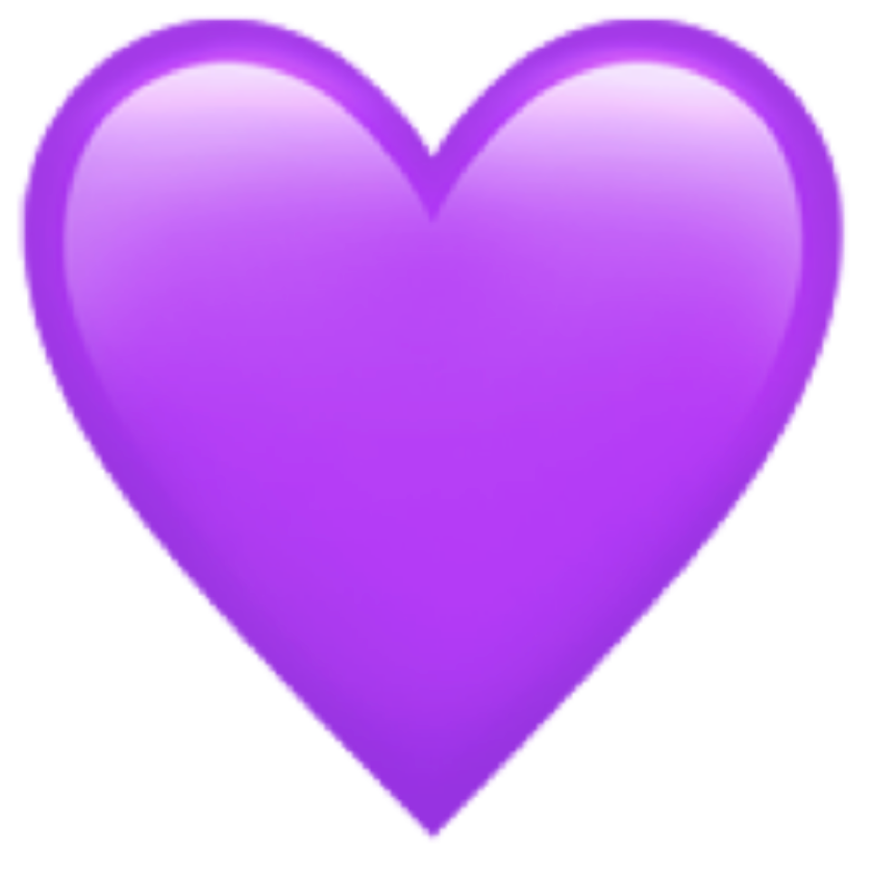}}{\strut}~}
\def\speakinghead{\scalerel*{\includegraphics{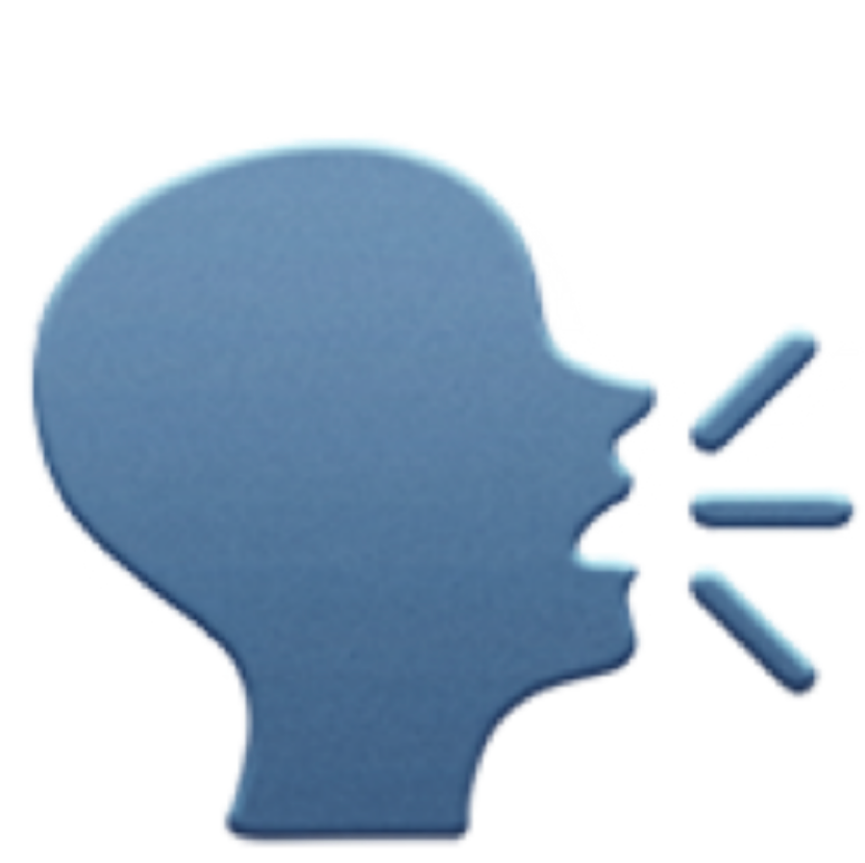}}{\strut}~}

\usepackage{microtype}

\usepackage{inconsolata}

\usepackage{graphicx}

\usepackage{booktabs}
\usepackage{verbatim}
\usepackage{subcaption}  
\usepackage{csquotes}

\usepackage{lscape}
\usepackage{enumitem}
\setlist[itemize]{noitemsep}
\usepackage{titletoc}

\title{When Detection Fails: The Power of Fine-Tuned Models \\to Generate Human-Like Social Media Text}

 \author{Hillary Dawkins \and Kathleen C. Fraser \and Svetlana Kiritchenko \\
         National Research Council Canada, Ottawa, Canada \\
         \small{\texttt{\{hillary.dawkins, kathleen.fraser, svetlana.kiritchenko\}@nrc-cnrc.gc.ca}}}

\begin{document}
\maketitle
\begin{abstract}

Detecting AI-generated text is a difficult problem to begin with; detecting AI-generated text on \textit{social media} is made even more difficult due to the short text length and informal, idiosyncratic language of the internet. It is nonetheless important to tackle this problem, as social media represents a significant attack vector in online influence campaigns, which may be bolstered through the use of mass-produced AI-generated posts supporting (or opposing) particular policies, decisions, or events. We approach this problem with the mindset and resources of a reasonably sophisticated threat actor, and create a dataset of 505,159 AI-generated social media posts from a combination of open-source, closed-source, and fine-tuned LLMs, covering 11 different controversial topics. We show that while the posts can be detected under typical ``research'' assumptions about knowledge of and access to the generating models, under the more realistic assumption that an attacker will not release their fine-tuned model to the public, detectability drops dramatically. This result is confirmed with a human study. 
Ablation experiments highlight the vulnerability of various detection algorithms to fine-tuned LLMs.
This result has implications across all detection domains, since fine-tuning is a generally applicable and realistic LLM use case. 

\end{abstract}

\section{Introduction}

Large language models (LLMs) are able to produce increasingly complex and fluent output, creating an information environment where humans can no longer reliably distinguish between text that was written by other humans, and that which was generated by LLMs. This increases our vulnerability to various tactics of opinion manipulation, particularly in online spaces such as social media. These tactics include disinformation and misinformation campaigns, as well as astroturfing and ``flooding the zone'' attacks. While growing research effort has focused generally on the development of methods for detecting  AI-generated text (AIGT), very little work has focused on the pressing question of AIGT detection in the social media domain. Furthermore, much of the research has relied on assumptions that are highly unlikely to hold in real-world contexts, such as the assumption that the defender has knowledge of (or even access to) the LLM that was used to generate the text they are trying to detect.

In the current work, we are particularly motivated by the 
concept of \textit{astroturfing}, or the artificial creation of an impression of widespread support for (or opposition to) a product, policy, or concept \cite{chan2024online}. By flooding the information space with posts advocating a particular stance on some issue, AI can be used to influence and change people's beliefs by exploiting their natural tendency toward ``crowd mentality'' and desire for social acceptance. Crucially, astroturfing is distinct from disinformation, in that the sentiment being expressed is not necessarily \textit{false}. This means that traditional methods of disinformation detection are not effective against such an attack. Rather, the deception involved in astroturfing is the \textit{inauthenticity of the apparent crowd} \cite{chan2024online}. Therefore, the ability to detect whether a social media post on some topic originates from a real human or from AI becomes an important signal for authenticity determination.

Our work attempts to make realistic assumptions about the motivations and resources available to both \textit{attackers} (those spreading the AIGT) and \textit{defenders} (those detecting the AIGT) in this scenario. In contrast to previous work, which 
sampled social media posts on a broad variety of everyday topics \cite{macko2024multisocial}, we assume that both attackers and defenders will be more motivated to participate in/monitor online discussions on controversial or political topics. We assume that the attacker will have a particular stance they want to communicate on a given topic, rather than simply summarizing or paraphrasing existing posts.  We also assume they will take reasonable, low-cost steps to produce realistic text (such as fine-tuning the model to mimic the intended social media style), and that they will not release any fine-tuned models for the defender to access. 
While it is reasonable to assume that an actor will attempt to mimic the intended distribution through fine-tuning, we do not make any further assumptions that the attacker is actively trying to evade detection, which may require more sophisticated knowledge of modern detection schemes. 



Therefore, we examine four related research questions: 

\begin{itemize}[itemsep=1pt, leftmargin=5.5mm]
    \item \textbf{RQ1: Can today's LLMs produce realistic social media style text?} We explore 
    prompting base models as well as models fine-tuned on human-written texts, and conduct linguistic analysis to compare the generated text with human text. We find that the output from fine-tuned models is more similar to human text than from the base models (Section \ref{sec:linguistic_results}). 
    \item \textbf{RQ2: Assuming a more realistic \textit{attack scenario}, do different generation strategies result in AIGT that is intrinsically harder to detect?}
    We generate text using different prompting conditions, with four base models and four fine-tuned models, and compare the classification accuracies of eight detection methods using idealized in-domain train-test classification methodology. We find that a supervised RoBERTa-based classifier can detect AIGT from the base models with up to 99\% accuracy, but this figure drops significantly for AIGT from the fine-tuned models (Section \ref{sec:detectability_results}).
    \item \textbf{RQ3: Assuming a more realistic \textit{defense scenario}, are detection methods still robust if they cannot access the generating model?} Many research studies assume at least black-box access to the generating model.
    Relaxing this assumption, we find that 
    the fine-tuned models are particularly hard to detect using off-the-shelf methods (Section \ref{sec:off-the-shelf-results}).
    Ablation studies reveal that access to closely related fine-tuned models is not a viable strategy using the tested detectors (Sections \ref{sec:viability_metrics}, \ref{sec:viability_plm}).
    \item \textbf{RQ4: Can human readers detect AI-generated texts?} In particular, we address the potential concern that fine-tuned models might generate text that automated methods cannot easily detect, but that human annotators can. We run a human study with 250 participants and find that texts from fine-tuned models are essentially undetectable to human readers (54\% detection accuracy) (Section \ref{sec:human_results}).
\end{itemize}

In addition to these research contributions, we also make our dataset of 505,159 English-language AI-generated texts available to other researchers. 
\section{Background}

Here, we briefly summarize the state-of-the-art NLP approaches to detecting AIGT, along with their strengths and weaknesses; more in-depth background is available in recent survey papers \cite{crothers2023machine,ghosal2023towards,uchendu2023survey,yang2023survey,fraser2024detecting,tang2023science,wu2023survey}.

One category of detection methods is \textit{metric-based} approaches that rely on detecting the statistical regularities of AIGT resulting from the generation process.
These include measures of the log-likelihood or rank of each token, with respect to the model's probability distribution over tokens at each step, or measures of the overall entropy or perplexity of the text \cite{gehrmann-etal-2019-gltr, su-etal-2023-detectllm}. The DetectGPT algorithm \cite{mitchell2023detectgpt} extends these ideas to estimate the local curvature of the model's log probability function,
laying the groundwork for Fast-DetectGPT
\cite{baofast}. These are all so-called ``zero-shot'' methods that do not require training data (although in practice, some in-domain data is needed to calibrate the decision threshold). They also rely on information about the probability distribution of the generating model, which ideally 
is obtained through 
white-box access to that model. 

Other methods use probability values from open-source LLMs as proxies for the unknown true probability distribution of the generating model \cite{li2023origin,wang2023seqxgpt,verma2024ghostbuster}.
The Binoculars method \cite{hans2024spotting} combines the benefits of black-box detection and ``zero-shot'' classification by measuring cross-entropy between two open-source models as a single feature to distinguish AIGT from human-written text.


In contrast to the feature-engineering methods described above, another broad set of AIGT-detectors involves fine-tuning pretrained language models (PLMs), such as  RoBERTa, for the classification task. Such methods are limited by the fact that they require larger quantities of labeled training data, and classifiers trained on older LLMs do not generalize to text from newer LLMs \cite{ghosal2023towards}. However, \citet{li-etal-2024-mage} found that, compared to metric-based methods, a fine-tuned PLM detector performed better when tested on unseen domains and unseen generating models. 

Very little work has focused specifically on social media text \cite{fagni2021tweepfake,kumarage2023stylometric,shao-etal-2024-natural}. One recent exception is the MultiSocial dataset, comprising 472,097 texts 
from seven multilingual LLMs, and covering five different social media platforms \cite{macko2024multisocial}. This dataset represents a valuable contribution to an understudied field. 
However, the data generation 
strategy involved broadly sampling social media posts and then generating AIGT paraphrases of the posts,
which does not accurately represent known threat models of AI-generated content on social media \cite{crothers2023machine,devereaux2025mockingird}. Here, we choose specific topics more likely to be targeted by influence campaigns, compare different methods of prompting to generate the AIGT, and explore the use of fine-tuned generating models. 


\section{Methods}

\subsection{Human-Written Data Collection}
\label{sec:human_data}


We restrict our data collection to X/Twitter posts, due to the wide availability of academic datasets collected through the Researcher API prior to the introduction of the paywall.
All human-written data is sourced from existing datasets, as outlined in Appendix Table~\ref{tab:human_data}, and restricted to posts written before March 2022, to minimize the chances of AIGT contaminating the sample. Eleven ``controversial'' topics are chosen that contain emotional language, divisive opinions, and misinformation, namely: climate change, abortion, feminism, refugees and migrants, data privacy, COVID-19, Women's March, MAGA, \#MeToo, Brexit, and the war in Ukraine.  We sample 1000 tweets per topic, for a total of 11,000 human-written tweets. Where possible, multiple sources are used for each topic to promote variety and mitigate bias in the human data distribution. Further information about the human-written data is available in Appendix~\ref{app:human}.

\subsection{Generative AI Models}
\label{sec:models}

We consider two broad model families in this work: OpenAI's GPT models (closed-source), and Meta's Llama family of models (open-source). Within these broad categories, we consider both larger and smaller models, to assess the effect of model size on generation quality. 
Specifically, we experiment with GPT-4o (version: 2024-08-06), GPT-4o-mini (version: 2024-07-18), Llama-3-8B-Instruct, 
and Llama-3.2-1B-Instruct.

\subsection{AIGT Generation Strategies}
\label{sec:generation}

Here we provide an overview of the data generation methods. Additional details, including exact prompts, parameter settings, and costs can be found in Appendix~\ref{app:generation}.

\subsubsection{Paraphrase}

To enable comparison with existing datasets, we first generate AIGT paraphrases of the human-written content. Specifically, we follow the generation procedure outlined by \citet{macko2024multisocial} of paraphrasing the original content iteratively three times, 
with the intention that each iteration of paraphrasing perturbs the text further from the original. 

\subsubsection{Generate From Example}

The iterative paraphrasing approach is unlikely to be used in an astroturfing campaign,
since it requires three API calls to generate a single tweet.
An alternative 
generation strategy 
for producing content at scale 
is to use the human-written tweet as an exemplar, and ask AI to create \textit{multiple different} tweets that convey the same meaning. 
We prompt the LLMs to generate 10 variations given a single human-written tweet.

\subsubsection{Generate From Topic}

Both of the previous methods assume the availability of an existing human-written tweet to use as an input.
We believe that most research in AIGT detection generates such ``pairs'' of human-AIGT samples not based on any evidence that this is how LLMs are actually used in practice, but rather due to the need to control for topic bias in the detection experiments. 
Since we agree with this methodological concern, but also aim to generate AIGT under a more realistic scenario, we propose a new method for deriving human-AIGT topic-matched pairs that does not rely on paraphrasing.

First, the main topic and stance are extracted for each human-written tweet by asking GPT-4o ``What is the main topic of this tweet, and what stance does the author take?'' The resulting description (denoted as \verb|Topic|) summarizes the content of the human-written tweet succinctly (see Appendix \ref{app:gen_from_topic} for an example). Second, each LLM is instructed to express the \verb|Topic|: ``Write a tweet in casual, social media style based on the following description: <\verb|Topic|>''.

We claim that this prompt, which asks for open-ended generation on a particular topic and expressing a particular stance, is a much closer approximation to how AIGT would be generated by attackers engaged in an online influence campaign.

\subsubsection{Fine-tuning}
\label{sec:fine_tuning}

To our knowledge, none of the previous research on AIGT detection includes data generated from fine-tuned models. However, even moderately-sophisticated actors would be able to develop a fine-tuned model using readily available information online, and cheap access to hardware through subscription services like Google Colab. Here, we adapt the LLMs to the language of social media by fine-tuning on a set of random-topic tweets (collected in 2019 and completely separate from the human-written tweets in the current dataset). We compare two conditions: fine-tuning on a `small' subset (800 training samples)
and a `large' subset (2000 training samples),

Base models GPT-4o and GPT-4o-mini were fine-tuned using OpenAI's online fine-tuning interface, which makes it very easy for non-technical users to create fine-tuned models, although exact methodological details are unknown to the user. Base models Llama-3-8B-Instruct and Llama-3.2-1B-Instruct were fine-tuned using QLoRA \cite{qlora, lora} (4-bit quantization, with an adapter rank of 64) via supervised fine-tuning (SFT) in a Google Colab environment. In all cases, fine-tuning proceeded for 5 epochs.   
The estimated costs and times for fine-tuning each model are given in Table~\ref{tab:finetuned_models}, but range between 3 cents to \$26 (USD) and 10 minutes to 3 hours, demonstrating how affordable and practical this process is. 
More details about the fine-tuning procedure, including the training sample template, are provided in Appendix \ref{app:data_gen:fine-tuning}. 
Examples of AI-generated text using each prompting strategy are provided in Table \ref{tab:example_outputs}.

\subsection{Linguistic Analysis}
To investigate linguistic differences in texts produced by different LLMs and by human users, we follow \citet{reinhart2024llms} and \citet{sardinha2024ai} and consider the set of linguistic features proposed by Douglas Biber, as implemented in the BiberPy Python package.\footnote{\url{https://github.com/ssharoff/biberpy}} Noting that many of those features may be more likely to occur in longer, more formal texts, we also examine a set of features that we expect to be more relevant in the context of social media: the number of `@' mentions, links, and emojis; 
the ratio of uppercase-to-lowercase characters; 
and the presence of offensive language\footnote{\url{https://www.cs.cmu.edu/~biglou/resources/}} and typos/nonstandard spellings\footnote{\url{https://pypi.org/project/pyspellchecker}}. 

\subsection{AIGT Detection Methods}

For the classification experiments, we include several existing metric-based detection methods: Log-Likelihood, Entropy, Rank, Log-Rank, the Log-Likelihood-Log-Rank Ratio \cite{su-etal-2023-detectllm}, and Fast-DetectGPT \cite{baofast}. These methods all use probability information and therefore require either white-box access to the generating model itself, or to some proxy measurement model.
We also consider Binoculars \cite{hans2024spotting}, which does not assume white-box access to the generating model. All of these metric-based models have decision thresholds which are ideally calibrated on in-distribution data, when available. We also evaluate a PLM detector (OpenAI's RoBERTa-based detector\footnote{\url{https://huggingface.co/openai-community/roberta-base-openai-detector}}), which can be used off-the-shelf or further fine-tuned on in-domain data. Finally, we include two other off-the-shelf methods: ChatGPT-Detector\footnote{\url{https://huggingface.co/Hello-SimpleAI/chatgpt-detector-roberta}} 
\cite{guo2023close} and GPTZero, a commercial offering.\footnote{\url{https://gptzero.me/}} Classification experiments were built upon an existing detection suite MGTBench
\cite{HSCBZ24_MGTB}.


We apply these detection methods in several scenarios, including the idealized case of having full access to the generating model and the training data, an off-the-shelf case with no knowledge of the generating model or access to training data, and various intermediate scenarios.
Note that we distinguish between the \textbf{generating model} (the LLM that generated the AIGT) and the \textbf{measurement model} (the LLM that provides the word probability estimates, where needed). 
The word `model' could also refer to a detection model; for clarity, we refer to such a model as a \textbf{detector}. 

\subsection{Human Study}

Studies have indicated that human annotators can no longer reliably distinguish between AIGT and human-written text \cite{liu2023check,sarvazyan2023overview}. However, our data differs from the data used in those studies in two important respects: (1) it is social media-style text, rather than academic writing, news, or reviews, and (2) we consider text from fine-tuned LLMs, and there is a chance that the fine-tuning process might introduce some artifacts that make the resulting text \textit{more} identifiable to human readers, regardless of the performance of the detection algorithms. Therefore, to confirm our hypothesis that text generated from fine-tuned models is harder to detect for both humans and machines, we run an online human annotation study. Full methodological details are available in the Appendix~\ref{app:human_study}, but in summary: we recruited 250 human participants on Amazon Mechanical Turk to categorize tweets as either AI-generated or human-written. In two separate studies, AI-generated tweets were obtained from either the base GPT4o model (3,640 human annotations) or the fine-tuned (large sample) GPT4o model (3,360 human annotations), and detection accuracy is compared across the two conditions.  
Ethics approval for this study was obtained from the Research Ethics Board of the National Research Council of Canada.

\section{Results}

\subsection{Linguistic Analysis}
\label{sec:linguistic_results}

The purpose of our linguistic analysis is to determine whether the texts that are generated by LLMs are significantly different from those written by humans.  To determine statistical significance, we use a nonparametric Mann-Whitney U-test due to the violation of the assumption of normality. However, due to the large number of samples (11,000 for each class), even very small differences are found to be highly significant.\footnote{All of the features reported in Table~\ref{tab:effect_size} show statistically significant differences ($p << 0.05$) between human and AIGT texts, with the exception of `Possibility modals' in the Human vs Fine-Tuned case.} Thus, rather than focusing on $p$-values, we report the effect size, as measured by the Rank-Biserial correlation.\footnote{\url{https://search.r-project.org/CRAN/refmans/effectsize/html/rank_biserial.html}} Table~\ref{tab:effect_size} reports the effect size of the difference for the linguistic features of the base GPT-4o model and the corresponding fine-tuned model, each compared to human-written tweets. 
The effect sizes can be interpreted as the difference in the proportion of samples in each group that rank higher in each feature, ranging from $-1$ to $+1$. For example, if \textit{all} text samples generated by GPT-4o contain more emojis than \textit{all} human-written text samples, the effect size would be $+1$ for the ``Emojis'' feature. Features for which there is a negligible difference ($|R| < 0.1$) between the human-written and AI-written texts are indicated in green. Other colors indicate a more meaningful effect size. 
The full results for all four LLMs are in Appendix \ref{app:linguistic_analysis}, along with additional analyses. In general, we observe the following trends: 

\begin{itemize}[itemsep=1pt, leftmargin=5.5mm]
    \item Base models tend to be verbose, generating tweets that are longer than human tweets (see also Appendix Fig.~\ref{fig:linguistic_dist}, top row). 
    \item The base models tend to have a lower type-token ratio (TTR) than human-generated text, indicating more repetitions and lower lexical diversity (see Fig.~\ref{fig:linguistic_dist}, middle row).
    \item Base models do not spontaneously generate certain characteristics of social media text, such as @mentions and links, but tend to over-generate hashtags (see Fig.~\ref{fig:linguistic_dist}, bottom row) and emojis.
    \item Certain linguistic features are significant across all four models, including the presence of more adverbs, first-person pronouns, private verbs (verbs expressing internal state), 
    and contractions compared to human-written text.
    \item The fine-tuned models show \textit{smaller} effect size differences than the base models 
    on almost all linguistic features. 
\end{itemize}

On the basis of this analysis, we conclude that \textbf{text output by the fine-tuned models is more similar to real human-written text on social media (RQ1)}, and we therefore expect it to be more difficult to detect than text from the base LLMs.

\begin{table}[tbh]
    \footnotesize
\begin{tabular}{lrr}
\toprule
Feature & Human vs Base & Human vs FT \\
\midrule
@mentions & \cellcolor{orange!25}-0.47 & \cellcolor{yellow!25}0.11 \\
Links & \cellcolor{orange!25}-0.38 & \cellcolor{green!25}-0.07 \\
Hashtags & \cellcolor{yellow!25}0.26 & \cellcolor{yellow!25}-0.29 \\
Emojis & \cellcolor{red!25} 0.92 & \cellcolor{green!25}0.05 \\
Length (chars) & \cellcolor{red!25} 0.60 & \cellcolor{green!25}0.08 \\
Offensive & \cellcolor{yellow!25}-0.11 & \cellcolor{green!25}0.02 \\
Upper:lower ratio & \cellcolor{yellow!25}-0.15 & \cellcolor{yellow!25}-0.18 \\
\hline
Past verbs & \cellcolor{yellow!25}-0.10 & \cellcolor{green!25}0.02 \\
Present verbs & \cellcolor{yellow!25}0.24 & \cellcolor{green!25}0.05 \\
First pers. pron. & \cellcolor{yellow!25}0.21 & \cellcolor{green!25}-0.03 \\
Impersonal pron. & \cellcolor{yellow!25}0.24 & \cellcolor{green!25}0.02 \\
Indefinite pron. & \cellcolor{yellow!25}0.14 & \cellcolor{green!25}0.03 \\
Wh- questions & \cellcolor{yellow!25}0.14 & \cellcolor{green!25}0.02 \\
Nominalizations & \cellcolor{yellow!25}0.11 & \cellcolor{green!25}0.05 \\
Attributive adj. & \cellcolor{yellow!25}0.15 & \cellcolor{green!25}0.03 \\
Adverbs & \cellcolor{yellow!25} 0.26 & \cellcolor{green!25}0.05 \\
Private verbs & \cellcolor{yellow!25}0.25 & \cellcolor{green!25}0.06 \\
Possibility modals & \cellcolor{yellow!25}0.13 & \cellcolor{green!25}0.00 \\
Contractions & \cellcolor{orange!25}0.39 & \cellcolor{green!25}-0.02 \\
That deletion &\cellcolor{yellow!25} 0.18 & \cellcolor{green!25}0.03 \\

\bottomrule
\end{tabular}
\caption{Effect size (Rank-Biserial Correlation) of differences between human-written texts and texts generated by \textbf{GPT-4o} (base model and fine-tuned (FT)). Positive values indicate that the feature value is higher in AIGT than in human-written text. Effect size can be interpreted as: $|R| < 0.1$ - \colorbox{green!25}{no effect} (features with no effect in either group are omitted from the current table; see Appendix \ref{app:linguistic_analysis} for full table), $|R| > 0.1$ - \colorbox{yellow!25}{small effect}, $|R| > 0.3$ - \colorbox{orange!25}{medium effect}, $|R| > 0.5$ - \colorbox{red!25}{large effect}.}
\label{tab:effect_size}
\end{table}

\subsection{AIGT Detection}

\subsubsection{Detectability (Idealized Case)}
\label{sec:detectability_results}


\setlength{\tabcolsep}{5pt}
\begin{table*}[tbh]
    \footnotesize
\begin{tabular}{llrrrr | rrrr}
\toprule
& & \multicolumn{4}{c}{\textbf{Complete Knowledge (Idealized) Scenario}} & \multicolumn{4}{c}{\textbf{Off-the-Shelf Scenario}} \\
\midrule 
LLM & Prompt & LL & F-DGPT & Bino. & PLM  & Bino. & PLM & ChatGPT & GPTZero \\
 &  & &  & (calibr.) & (fine-tuned) &  &  & Detector &  \\
\midrule
Llama-3.2-1B & Para-1  &  0.792 &  0.711 & 0.653 & \textbf{0.931}   & 0.604 & 0.591 & 0.580 & \textbf{0.747} \\
Llama-3.2-1B & Para-2  &  0.815 &  0.698 & 0.666 & \textbf{0.971}   & 0.604 & 0.552 & 0.585 & \textbf{0.805} \\
Llama-3.2-1B & Para-3  &  0.809 &  0.672 & 0.669 & \textbf{0.988}   & 0.599 & 0.536 & 0.585 & \textbf{0.829} \\
Llama-3.2-1B & Gen-10  &  0.605 &  0.695 & 0.608 & \textbf{0.920}   & 0.596 & \textbf{0.668} & 0.546 & 0.661 \\
Llama-3-8B & Para-1  & 0.653  & 0.632 & 0.629 & \textbf{0.941}  & 0.589 & 0.536 & 0.557 & \textbf{0.749} \\
Llama-3-8B & Para-2  & 0.621 & 0.606 & 0.618 & \textbf{0.972}  & 0.569 & 0.536 & 0.561 & \textbf{0.799} \\
Llama-3-8B & Para-3  & 0.677 & 0.534 & 0.570 & \textbf{0.972} & 0.542 & 0.517 & 0.573 & \textbf{0.846} \\
Llama-3-8B & Gen-10  & 0.691 & 0.635 & 0.686 & \textbf{0.911}  & 0.655 & 0.533 & 0.572 &\textbf{ 0.751} \\
GPT-4o-mini & Para-1  & - & - & 0.538 & \textbf{0.903}  & 0.513 & 0.502 & 0.543 & \textbf{0.662} \\
GPT-4o-mini & Para-2  & - &  - & 0.539 & \textbf{0.891}  & 0.508 & 0.509 & 0.550 & \textbf{0.690} \\
GPT-4o-mini & Para-3  & - & - & 0.546 & \textbf{0.945} & 0.507 & 0.515 & 0.554 & \textbf{0.747} \\
GPT-4o-mini & Gen-10  & - &  - & 0.553 & \textbf{0.873} & 0.526 & 0.515 & 0.565 & \textbf{0.659} \\
GPT-4o & Para-1  & - &  - & 0.520 & \textbf{0.900}  & 0.498 & 0.505 & 0.548 & \textbf{0.687} \\
GPT-4o & Para-2  & - & - &  0.519 & \textbf{0.916} & 0.508 & 0.520 & 0.561 & \textbf{0.716} \\
GPT-4o & Para-3  & - & - & 0.523 & \textbf{0.940}  & 0.506 & 0.507 & 0.574 & \textbf{0.775} \\
GPT-4o & Gen-10  & - & - & 0.528 & \textbf{0.867} & 0.526 & 0.509 & 0.560 & \textbf{0.623} \\
\hline
\midrule
Llama-3.2-1B & Topic  & 0.905 &  0.800 & 0.728 & \textbf{0.990} & 0.645 & 0.512 & 0.547 & \textbf{0.822} \\
FT\_Llama-1B-small & Topic  & 0.687 &  0.767 & 0.567 & \textbf{0.857} &  0.533 & \textbf{0.582} & 0.547 & 0.496 \\
FT\_Llama-1B-large & Topic  & 0.708 &  0.711 & 0.559 & \textbf{0.838}  & 0.517 & \textbf{0.601} & 0.519 & 0.498 \\
\midrule
Llama-3-8B & Topic  & 0.928 & 0.692 & 0.788 & \textbf{0.992}  & 0.737 & 0.403 & 0.539 & \textbf{0.912 }\\
FT\_Llama-8B-small & Topic  & 0.694 &  0.732 & 0.641 & \textbf{0.837} & \textbf{0.610} & 0.534 & 0.571 & 0.517 \\
FT\_Llama-8B-large & Topic  & 0.653 &  0.720 & 0.608 & \textbf{0.803} & \textbf{0.565} & 0.523 & 0.517 & 0.499 \\
\midrule
GPT4o-mini & Topic  & - & - & 0.758 & \textbf{0.997} & 0.682 & 0.404 & 0.583 & \textbf{0.971 }\\
FT\_GPT4o-mini-small & Topic  & - & - & 0.585 & \textbf{0.782}  & \textbf{0.581 }& 0.535 & 0.546 & 0.512 \\
FT\_GPT4o-mini-large & Topic  & - & - & 0.545 & \textbf{0.741}  & 0.513 & \textbf{0.518 }& 0.513 & 0.499 \\
\midrule
GPT4o & Topic  & - & - &  0.650 & \textbf{0.999} & 0.569 & 0.380 & 0.551 & \textbf{0.945} \\
FT\_GPT4o-small & Topic  & - & - &  0.509 & \textbf{0.723}  & 0.515 & \textbf{0.533} & 0.509 & 0.501 \\
FT\_GPT4o-large & Topic  & - & - & 0.491 & \textbf{0.716} & 0.499 & \textbf{0.516} & 0.510 & 0.494 \\
\bottomrule
\end{tabular}
\caption{\textit{On the left}, the \textbf{detectability} of the AIGT using idealized detectors (reported by accuracy). `FT' stands for a fine-tuned generator. Log-Likelihood (\textbf{LL}) and Fast-DetectGPT (\textbf{F-DGPT}) use the generating model as the measurement model, and these methods plus Binoculars (\textbf{Bino.}) have decision thresholds calibrated using in-distribution data. The pre-trained classifier (\textbf{PLM}) is further fine-tuned on the data distribution to detect. \textit{On the right}, accuracy of the detectors in the off-the-shelf scenario. Here, Binoculars uses the default decision threshold and the PLM is not further fine-tuned. The best detection accuracy for each generator, in each scenario, is in bold.
}
\label{tab:detectability_combined}
\end{table*}
\setlength{\tabcolsep}{6pt}

Firstly, we examine how \textit{intrinsically detectable} the AIGT is by measuring the classification accuracy under ideal conditions. 
In this setup, all measurement-model dependent methods use the generating model as the measurement model (i.e., white-box access to a known generator is assumed) where available. Additionally, all metric-based detectors have decision thresholds calibrated using in-distribution data, and the pre-trained RoBERTa detector is further fine-tuned on in-distribution data. 
Although these conditions do not represent a real-world scenario, they reveal which LLMs and generation methods most successfully mimic the human-written data distribution. 

The classification accuracies for a subset of the detectors are shown on the left side of Table \ref{tab:detectability_combined}. Full results for all detectors are provided in the Appendix, Table \ref{tab:detectability_full}. Each dataset is balanced between the human-written and AI-generated classes; a full description of the evaluation datasets, including post-generation processing details, are provided in Appendix \ref{app:data_gen:processing}. Results showing true positive rate at a specific false positive rate, another useful metric in AIGT detection \cite{tufts2024practical}, can also be found in Appendix \ref{app:AIGT_detection_results}. 
As expected, the fine-tuned PLM is the strongest detector under this complete-knowledge scenario due to its ability to pick up on \textit{any} differences in the human and AI-generated data distributions, which are not fully captured by the predefined metric-based classifiers. 


Focusing on the PLM results then, 
we observe that iterative paraphrasing in fact increases detectability 
(higher accuracy for Para-3 than Para-1), and that the ``generate from example'' prompt yields slightly more evasive text than paraphrasing (lower accuracy for Gen-10 than Para-3). 
Both paraphrasing and generating from example are more evasive than asking for a tweet on a given topic \textit{without} giving a human reference (base model with ``Topic'' prompt). 
This suggests that the inclusion of a human-written example in the prompt may be an important consideration for generating more human-like text.

However, we observe that the fine-tuned generators are the most successful at producing human-like text, in terms of detectability, even without the resource of a human reference at generation time. Using larger models and larger amounts of fine-tuning data increases the evasiveness. In particular, a fine-tuned GTP4o model leads to the lowest detection accuracy at 71.6\%, down from nearly complete detectability (99.9\%) using the base model with the same prompting strategy. Therefore, 
\textbf{while prompt strategy does play a role in how detectable the output text is, the biggest drop in detectability occurs with fine-tuning (RQ2).}

\subsubsection{Off-the-Shelf Performance}
\label{sec:off-the-shelf-results}

In contrast to detectability under the most idealized detection conditions, we now turn our attention to the practical performance of off-the-shelf detectors. That is, we work under the assumption that no additional resources (either specific measurement models or calibration/training data) are available to power the detection methods. Off-the-shelf detectors include pre-trained classifiers, measurement model-agnostic metric-based classifiers with threshold values supplied by the developer, and commercially available detectors. Here, we consider four off-the-shelf detectors. Binoculars is implemented with the author-suggested decision threshold,\footnote{\url{https://github.com/ahans30/Binoculars}} rather than calibrating on in-distribution data as in the previous section. We consider two PLM detectors: OpenAI's detector and the ChatGPT detector; 
both use RoBERTa fine-tuned for AIGT classification using external GPT-family datasets. GPTZero is a commercial product, accessed through the API.

Detection accuracies are reported on the right side of Table \ref{tab:detectability_combined}.
In general, existing off-the-shelf detectors are insufficient solutions for classifying short, social-media style texts.
GPTZero performs well on base model generations, especially when prompts do not include a human reference (``Topic prompt'' rows in the bottom half of Table~\ref{tab:detectability_combined}). However, fine-tuning the base models consistently decreases GPTZero's detection accuracy to random-baseline levels. In the fine-tuned case, other detectors perform marginally better than GPTZero, but still with a maximum accuracy of 61\% (Bino.). This indicates that \textbf{off-the-shelf methods, which do not require knowledge of the generating LLM, are not reliable detectors of 
social media texts generated by fine-tuned models (RQ3).}

\subsubsection{Viability of Metric-based Detectors}
\label{sec:viability_metrics}

The previous two sections establish opposite ends of the classification results spectrum: the complete-knowledge scenario, and the no-knowledge scenario.
Between these two extremes, existing methods can be augmented with some additional resources, but assumptions of complete knowledge should be relaxed to assess more realistic scenarios. 
For both metric-based and PLM-based detectors, we assess the viability of the methods by first \textit{minimally} reducing the resources from the idealized scenario and observing their performance. 

For metric-based detectors, required resources include access to a measurement model to calculate the metric, and 
calibration data to set a decision threshold. 
Here we 
relax the assumption that we can access the generating model for use as the measurement model. 
Note that this is especially relevant to the case of fine-tuned generators because the fine-tuning procedure and the fine-tuned model are most likely kept private. 
Though we may not have access to the exact generating model, we might assume access to 
the base model or a similarly fine-tuned model.

Under ideal conditions, Fast-DetectGPT is the best-performing metric-based detector on the fine-tuned generators (Table \ref{tab:detectability_full}). 
Its performance matrix of generating model $\times$ measurement model is shown in Table \ref{tab:Fast-detectgpt_auc_bino}. We report classification performance using threshold-free evaluation (AUROC) to remove dependence on any calibration data.
GPT-2 is included as a baseline measurement model, and Binoculars is included as a measurement-model-agnostic comparison. 

Surprisingly, very closely related fine-tuned models are just as weak, if not weaker, than base models in the measurement model role.  
For Llama-3.2-1B and Llama-3-8B, it is better to use the base Llama models as the measurement model than any of the `sister' fine-tuned models (built from the same base model using the same procedure, 
just with more/fewer examples from the same fine-tuning dataset).
The same observation holds for the other measurement model-dependent detectors (see Appendix \ref{app:AIGT_detection_results} for
Log-Likelihood (\ref{tab:ll_auc}), Entropy (\ref{tab:entropy_auc}), Rank (\ref{tab:rank_auc}), Log-Rank (\ref{tab:log_rank_auc}), and LLR (\ref{tab:LLR_auc})),
implying that the measurement-model dependent detectors are not robust to minor changes in the fine-tuned measurement models. 
For the closed-source fine-tuned models, 
detection is not feasible using any of the tested 
measurement models, despite being fine-tuned on the same data. 



\begin{table*}[tbhp]
\centering
    \footnotesize
\begin{tabular}{lr|rrr|rrr|r}
\toprule
 & \multicolumn{7}{c|}{Fast-DetectGPT by measurement model} & Bino. \\
Generating LLM  & GPT-2 & Llama-1B & FT-1B-sm & FT-1B-lg & Llama-8B & FT-8B-sm & FT-8B-lg  & \\
\midrule
Llama-3.2-1B & \underline{0.851} &\cellcolor{green!25} \textbf{0.879} & \cellcolor{yellow!25}0.671 &\cellcolor{yellow!25} 0.661 & \cellcolor{orange!25}0.531 & \cellcolor{orange!25}0.615 & \cellcolor{orange!25}0.507  & 0.803\\
FT\_Llama-1B-small & 0.586 & \cellcolor{yellow!25}\underline{0.720} & \cellcolor{green!25} \textbf{0.815} & \cellcolor{yellow!25}0.690 & \cellcolor{orange!25}0.605 & \cellcolor{orange!25}0.568 & \cellcolor{orange!25}0.553  & 0.592\\
FT\_Llama-1B-large & 0.555 & \cellcolor{yellow!25}\underline{0.702} & \cellcolor{yellow!25}0.640 & \cellcolor{green!25} \textbf{0.772} & \cellcolor{orange!25}0.584 & \cellcolor{orange!25}0.501 & \cellcolor{orange!25}0.549  & 0.577\\
\midrule
Llama-3-8B & \underline{0.855} & \cellcolor{orange!25}0.717 & \cellcolor{orange!25}0.584 & \cellcolor{orange!25}0.592 & \cellcolor{green!25} 0.750 & \cellcolor{yellow!25}0.554 & \cellcolor{yellow!25}0.659  & \textbf{0.857}\\
FT\_Llama-8B-small & 0.666 & \cellcolor{orange!25}0.702 & \cellcolor{orange!25}0.713 & \cellcolor{orange!25}0.683 & \cellcolor{yellow!25}0.722 & \cellcolor{green!25} \textbf{0.810} & \cellcolor{yellow!25}\underline{0.738}  & 0.691\\
FT\_Llama-8B-large & 0.592 & \cellcolor{orange!25}0.650 & \cellcolor{orange!25}0.626 & \cellcolor{orange!25}0.661 & \cellcolor{yellow!25}\underline{0.689} & \cellcolor{yellow!25}0.672 & \cellcolor{green!25} \textbf{0.796}  & 0.637\\
\midrule
GPT4o-mini & 0.649 & \underline{0.716} & 0.556 & 0.535 & 0.657 & 0.639 & 0.501  & \textbf{0.822}\\
FT\_GPT4o-mini-small & 0.610 & 0.630 & \cellcolor{purple!25}\textbf{0.652} & \underline{0.647} & 0.624 & \cellcolor{purple!25}0.640 & 0.644  & 0.623\\
FT\_GPT4o-mini-large & 0.532 & \textbf{0.560} & 0.509 & \cellcolor{purple!25}0.525 & 0.551 & 0.514 & \cellcolor{purple!25}0.547  & \underline{0.556}\\
\midrule
GPT4o & 0.631 & \textbf{0.711} & 0.491 & 0.528 & 0.670 & 0.658 & 0.560  & \underline{0.699}\\
FT\_GPT4o-small & 0.501 & 0.534 & \cellcolor{purple!25} \textbf{0.560} & \underline{0.550} & 0.528 & \cellcolor{purple!25}0.543 & 0.548  & 0.513\\
FT\_GPT4o-large & 0.507 & \textbf{0.534} & \underline{0.527} & \cellcolor{purple!25}0.509 & 0.514 & 0.524 & \cellcolor{purple!25}0.491  & 0.495\\
\bottomrule
\end{tabular}
\caption{\textbf{Viability of metric-based detectors}: 
Classification potential (AUROC) of  Fast-DetectGPT and Binoculars. The Fast-DetectGPT matrix shows measurement models (columns) against generating models (rows). 
Measurement models may be equal to the generating model (green), share a common base model (yellow), share a common model family (orange), or share a common fine-tuning dataset (purple).
For each generating LLM, the best-performing detector is in bold and the second-best is underlined.}
\label{tab:Fast-detectgpt_auc_bino}
\end{table*}

\begin{table*}[tbhp]
\centering
    \footnotesize
\begin{tabular}{lrrr|rrr}
\toprule
 & \multicolumn{6}{c}{PLM-based detector by training dataset}  \\
Generating LLM  & Llama-1B & FT-1B-sm & FT-1B-lg & Llama-8B & FT-8B-sm & FT-8B-lg \\
\midrule
Llama-3.2-1B & \cellcolor{green!25}\textbf{0.981} & \cellcolor{yellow!25}0.748 & \cellcolor{yellow!25}0.773 & \cellcolor{orange!25}\underline{0.959} &\cellcolor{orange!25} 0.723 & \cellcolor{orange!25}0.745 \\
FT\_Llama-1B-small & \cellcolor{yellow!25}0.561 & \cellcolor{green!25}\textbf{0.864} &\cellcolor{yellow!25} \underline{0.782} & \cellcolor{orange!25}0.509 & \cellcolor{orange!25}0.767 & \cellcolor{orange!25}0.765 \\
FT\_Llama-1B-large & \cellcolor{yellow!25}0.559 & \cellcolor{yellow!25}0.749 & \cellcolor{green!25}\textbf{0.824} & \cellcolor{orange!25}0.514 & \cellcolor{orange!25}0.707 &\cellcolor{orange!25} \underline{0.771} \\
\midrule
Llama-3-8B &\cellcolor{orange!25} \underline{0.981} &\cellcolor{orange!25} 0.653 &\cellcolor{orange!25} 0.682 & \cellcolor{green!25}\textbf{0.996} &\cellcolor{yellow!25} 0.730 &\cellcolor{yellow!25} 0.699 \\
FT\_Llama-8B-small & \cellcolor{orange!25}0.557 & \cellcolor{orange!25}0.758 &\cellcolor{orange!25} 0.716 & \cellcolor{yellow!25}0.510 & \cellcolor{green!25}\textbf{0.841} & \cellcolor{yellow!25}0.769 \\
FT\_Llama-8B-large & \cellcolor{orange!25}0.540 &\cellcolor{orange!25} 0.680 & \cellcolor{orange!25}\underline{0.729} & \cellcolor{yellow!25}0.505 & \cellcolor{yellow!25}0.712 &\cellcolor{green!25} \textbf{0.801} \\
\midrule
GPT4o-mini & 0.946 & 0.831 & 0.802 & 0.990 & 0.872 & 0.818 \\
FT\_GPT4o-mini-small & 0.533 & \cellcolor{purple!25}0.704 & 0.666 & 0.502 & \cellcolor{purple!25}0.730 & 0.717 \\
FT\_GPT4o-mini-large & 0.532 & 0.604 & \cellcolor{purple!25}0.657 & 0.509 & 0.604 &\cellcolor{purple!25} \underline{0.689} \\
\midrule
GPT4o & 0.953 & 0.757 & 0.727 & 0.978 & 0.815 & 0.757 \\
FT\_GPT4o-small & 0.517 & \cellcolor{purple!25}0.639 & 0.649 & 0.507 & \cellcolor{purple!25}0.659 & 0.676 \\
FT\_GPT4o-large & 0.514 & 0.605 & \cellcolor{purple!25}0.653 & 0.502 & 0.596 & \cellcolor{purple!25}0.670 \\
\hline
\midrule
 & GPT4o-mini & FT-mini-sm & FT-mini-lg & GPT4o & FT-GPT4o-sm & FT-GPT4o-lg  \\
\midrule
Llama-3.2-1B & 0.713 & 0.768 & 0.774 & 0.869 & 0.700 & 0.644 \\
FT\_Llama-1B-small & 0.516 & \cellcolor{purple!25}0.778 & 0.737 & 0.521 & \cellcolor{purple!25}0.774 & 0.734 \\
FT\_Llama-1B-large & 0.508 & 0.729 & \cellcolor{purple!25}0.760 & 0.512 & 0.738 & \cellcolor{purple!25}0.738 \\
\midrule
Llama-3-8B & 0.803 & 0.731 & 0.751 & 0.944 & 0.691 & 0.579 \\
FT\_Llama-8B-small & 0.510 & \cellcolor{purple!25}\underline{0.777} & 0.703 & 0.518 & \cellcolor{purple!25}0.754 & 0.677 \\
FT\_Llama-8B-large & 0.503 & 0.718 &\cellcolor{purple!25} 0.720 & 0.504 & 0.711 & \cellcolor{purple!25}0.682 \\
\midrule
GPT4o-mini & \cellcolor{green!25}\textbf{0.998} & \cellcolor{yellow!25}0.831 &\cellcolor{yellow!25} 0.864 & \cellcolor{orange!25}\underline{0.995} & \cellcolor{orange!25}0.812 & \cellcolor{orange!25}0.742 \\
FT\_GPT4o-mini-small & \cellcolor{yellow!25}0.510 & \cellcolor{green!25}\textbf{0.786} &\cellcolor{yellow!25} 0.707 & \cellcolor{orange!25}0.515 &\cellcolor{orange!25} \underline{0.752} &\cellcolor{orange!25} 0.681 \\
FT\_GPT4o-mini-large & \cellcolor{yellow!25}0.506 & \cellcolor{yellow!25}0.674 & \cellcolor{green!25}\textbf{0.745} &\cellcolor{orange!25} 0.508 & \cellcolor{orange!25}0.681 & \cellcolor{orange!25}0.654 \\
\midrule
GPT4o &\cellcolor{orange!25} \underline{0.996} &\cellcolor{orange!25} 0.801 & \cellcolor{orange!25}0.821 & \cellcolor{green!25}\textbf{0.996} & \cellcolor{yellow!25}0.782 & \cellcolor{yellow!25}0.679 \\
FT\_GPT4o-small & \cellcolor{orange!25}0.507 & \cellcolor{orange!25}\underline{0.706} & \cellcolor{orange!25}0.679 & \cellcolor{yellow!25}0.512 & \cellcolor{green!25}\textbf{0.751} &\cellcolor{yellow!25} 0.667 \\
FT\_GPT4o-large &\cellcolor{orange!25} 0.503 & \cellcolor{orange!25}0.656 &\cellcolor{orange!25} 0.684 & \cellcolor{yellow!25}0.505 & \cellcolor{yellow!25}\underline{0.694} & \cellcolor{green!25}\textbf{0.719} \\
\bottomrule
\end{tabular}
\caption{\textbf{Viability of PLM-based detectors}: Classification accuracy of PLM-based detectors when trained on data from different generating models. For each generating LLM (row), the best-performing detector (training set) is in bold and the second-best is underlined. 
Green indicates that the PLM has been fine-tuned on the exact data distribution to detect (i.e., the training set and evaluation set are generated by the same model). Yellow indicates that the training set has been generated by a model sharing the same base model. Outside of shared base models, orange indicates that the training set has been generated by a model within the same model family (Llama or GPT). Outside model families, purple indicates that the training set has been generated by a model that was fine-tuned on the same data as the generator.}
\label{tab:FT_PLM}
\end{table*}

\subsubsection{Viability of PLM-based Detectors}
\label{sec:viability_plm}

Similarly, we assess the viability of PLM-based detectors by relaxing the resource assumptions. 
For this type of classifier, the ideal resource is access to labeled training data from the generating model.
As a first step toward realistic resource assumptions, we assess training data generated by similar models, using the same prompts and human-written data. 

Here we use OpenAI's RoBERTa-based detector as the PLM, and further fine-tune the classifier on our various AIGT datasets.
The performance matrix (evaluation dataset $\times$ training dataset) is shown in Table \ref{tab:FT_PLM}. 
Similarly to the metric-based detectors, we find that working with imperfect knowledge of a fine-tuned generator is detrimental to PLM-based detector performance. 
We first note that the base models do not generate high-quality training data for detecting any fine-tuned model 
(accuracy between 50-56\%). 
However, even closely related fine-tuned models do not generate training data that is a good substitute for in-distribution training data. For example, the best detection accuracy for fine-tuned Llama-3.2-1B-small is 86.4\%, 
trained on in-distribution data. 
Using training data generated by fine-tuned Llama-3.2-1B-large causes the detection accuracy to drop to 78.2\%. Using other fine-tuned models 
results in comparable drops in detection accuracy. 
This observation contrasts with the detection accuracy of the base generators, where generalization across models is possible in some cases. 

We stress that all fine-tuned generators were fine-tuned with the same dataset (of varying sizes), using similar procedures. Furthermore, the PLM training sets were generated using the same prompting strategy and human data distribution. In other words, the resource relaxation here represents a very minimal step towards a truly realistic detection scenario, and even so, detection accuracy is significantly affected. 
Together with the previous section, we find that both branches of 
detection methods are quite brittle to extremely 
small perturbations in the fine-tuned model as a resource.
Thus, \textbf{detection is not feasible using these methods under the assumption that the exact fine-tuned generating model is private, even if one assumes access to closely related fine-tuned generators as a resource (RQ3).}

\subsection{Human Annotation Study}
\label{sec:human_results}

After filtering out likely unreliable annotations (see Appendix~\ref{app:human_study} for details), 
the average accuracy ($\pm$ 1 standard deviation) of human participants in determining the source of the tweets (AI-generated vs.\@ human-written)
is \textbf{61.2\%} ($\pm$ 1.68\%) for the base model and \textbf{53.9\%} ($\pm$ 0.82\%) for the fine-tuned model. 
The observed difference in the means is significant with a $p$-value of 0.0007. 
\textbf{This confirms that the texts generated by the fine-tuned model are harder for an average human to detect than the texts generated by the base LLM (RQ4).}

\section{Conclusion}

A realistically motivated attacker is likely to fine-tune a model for their specific style and use case, as it is cheap and easy to do so.
With minimal effort, time, and money, we produced fine-tuned generators that are capable of much more realistic social-media tweets, based on both linguistic features 
and detection accuracy, 
and verified through human annotations. 
Although 
motivated by the 
spread of AI content on social media, and the associated risks of astroturfing and 
influence campaigns, we stress that the main findings extend across all text domains. Indeed, fine-tuning models for style-specific content generation is a generally applicable method, and one that is likely already in use by many generative AI users -- calling into question whether existing methods of detecting AIGT are as effective in the real world as in the research lab.

\clearpage

\section{Limitations}

Our study focused exclusively on English. Future work should examine other languages, as well as the use of LLMs to create translated output for social media. Furthermore, we examined only two model families (OpenAI's GPT and Meta's Llama), while numerous other open- and closed-source options exist and may show different trends. 
In addition, we have experimented with limited sets of parameter values for model fine-tuning and prompt variants. The goal of this work is not to find the best way to automatically generate hard-to-detect social media texts, but rather to demonstrate that a relatively straightforward and cheap method of fine-tuning an LLM can produce a generator that is virtually undetectable in a real-world scenario. 

All of our human data were sourced from X/Twitter, due to the widespread existence of publicly available datasets from which to sample. The results are most likely to generalize to other platforms that are characterized by short posts, such as Threads or Bluesky. Text length is a known factor in AIGT detectability, and detectors may perform better on social media platforms that encourage longer posts, such as Reddit. All of the AIGT data was generated by us, using a small set of prompts, which may limit its generalizability and ecological validity. However, collecting AIGT from real online sources is highly challenging, since it entails first differentiating AI- and human-generated content -- i.e., the entire goal of the work itself \citep{cui2023said, sun2024we}.

In this study, we examine the challenges posed to detectors by the short, informal nature of social media writing, the impact of different prompting strategies, and the effect of fine-tuning to improve the realism of the generated text. Out-of-scope of the current analysis is the effect of 
\textit{adversarial attacks} aimed specifically at avoiding detection. Other research has found that detection accuracy can be significantly reduced by such attacks, including character- and word-level perturbations \cite{pu2023deepfake, wang2024stumbling, wu2024detectrl}, paraphrasing (as a defense, rather than a generation strategy) \cite{sadasivan2023can, krishna2024paraphrasing, masrour2025damage}, and prompt-tuning \cite{kumarage-etal-2023-reliable, shi2024red}. The effect of applying such methods on social-media text, before or after fine-tuning, is an open question for future research. 

We focus exclusively on the signals available in the generated content itself; however, real-world attempts to detect malign influence campaigns on social media should combine multiple sources of information, including characteristics of the user account (username, photo, posting frequency, etc.) as well as network characteristics (e.g., groups of accounts acting in a coordinated fashion) \cite{forrester2019propaganda, cresci2020decade}. Investigating the authenticity of images, videos, or linked content can also provide valuable information.

\section{Ethics Statement}

It is possible that the presented findings (namely, that fine-tuned models are more difficult to detect) could be used by malicious actors to create harder-to-detect AIGT. However, it is more likely the case that they are \textit{already doing so}. There are many easy-to-follow tutorials online explaining how to fine-tune open-source models and it would be naive to assume that they are being used only for benign purposes.  Rather, we suggest that as NLP researchers, we too often set up ``straw man'' detection experiments that are disconnected from how LLMs are used (and will be used) in the real world. 

We would also like to emphasize that producing text using generative AI models 
is not an inherently problematic issue that requires close monitoring on social media or other platforms. Generative AI can provide valuable assistance to users not proficient in the language of communication, helping them to write clear and effective messages. Furthermore, AIGT detectors may exhibit biases, for example, having higher false positive rates for certain demographic groups or people with different linguistic skills \cite{liang2023gpt}. 
Therefore, AIGT detection should be approached not as a goal in and of itself, but with a particular harmful outcome in mind (e.g., academic fraud, spread of misinformation, astroturfing), and with the acknowledgment that in most cases the detection of AI content is only one piece of evidence in a broader assessment.


\section*{Acknowledgements}
This project was funded by the Canadian Department of National Defence, Annex Number: ER-VAL-24-045. The authors thank Zachary Devereaux for insightful feedback and discussion.

\bibliography{custom}

\appendix

\startcontents[sections]
\printcontents[sections]{l}{1}{\setcounter{tocdepth}{2}}

\section{Human-Written Data Collection}
\label{app:human}

The sources of human-written data are summarized in Table~\ref{tab:human_data}. 
A handful of the sources documented specific search terms that were used to collect tweets through the Twitter Researcher API. The known collection tags are summarized in Table~\ref{tab:collection_tags}. Note that the SemEval-2016 source \cite{mohammad-etal-2016-semeval} for topics Feminism, Abortion, and Climate Change purposefully uses search terms that should encompass ``both sides'' of an issue (e.g., \#ProChoice and \#ProLife). Topics that do not have specific collection tags were sourced using alternative search methods. The F3 (CoAID) dataset \cite{cui2020coaid, lucas2023fighting} for COVID-19 uses news article titles as the search query where news articles were obtained by searching for topics ``COVID-19, coronavirus, pneumonia, flu (excluding Influenza A/B, bird flu and swine flu), lockdown, stay home, quarantine and ventilator''. The Brexit Polarity Tweets dataset was obtained by collecting tweets from Pro-Brexit and Anti-Brexit twitter accounts, as determined by the account bio. The MiDe22 dataset \cite{toraman-etal-2024-mide22} was a valuable resource for us in collecting tweets for the COVID-19, Climate Change, Refugees and Migrants, and Ukraine topics. This dataset contains tweets related to specific potential misinformation events, as detailed in Table~\ref{tab:MiDe_events}. By using a variety of events, each with their own specific collection tags, the bias in the human data for these topics is reduced. The remaining datasets do not have clear documentation about their collection practices. 
Where applicable, we further filter the datasets to contain the more opinionated tweets; we take the ``pro'' and ``anti'' subsets of the Twitter Climate Change Sentiment Dataset and the Brexit Polarity Tweets Dataset, excluding the ``neutral'' sentiment cases. 
Lastly, all datasets are filtered to contain only tweets written before March 2022 in English.

\setlength{\tabcolsep}{5pt}
\begin{table*}[tbh]
\centering
\footnotesize
\begin{tabular}{llllr}
\toprule
\textbf{Topic} &         \textbf{Source} &                \textbf{Year}& \textbf{License} & \textbf{$n$ Tweets}  \\
& & & (if specified) & \\
\midrule
COVID-19	&	\href{https://github.com/mickeymst/F3}{F3 (CoAID)}	&	May - Dec 2020 &	&	600	\\
	&	\href{https://github.com/metunlp/MiDe22/tree/main}{MiDe22} (10 events)	&	Oct 2021 - Feb 2022	& MIT	& 400	\\
Data Privacy	&	\href{https://www.kaggle.com/datasets/cerolacia/its-controversial/data?select=Enron.csv}{It's Controversial} 	&	2020	& CC0	&1000	\\
MAGA	&	\href{https://data.world/wendyhe/tweets-on-womensmarch-and-maga}{WomensMarch and MAGA Tweet Dataset}	&	2017	& CC0	&1000	\\
Women's March	&	\href{https://data.world/wendyhe/tweets-on-womensmarch-and-maga}{WomensMarch and MAGA Tweet Dataset}	&	2017	& CC0	&1000	\\
\#MeToo	&	\href{https://data.world/hamdan/tweets-with-emojis-metoo-2017-10-16}{Tweets With Emojis - \#MeToo}	&	Oct 16 2017	& CC BY-NC	&1000	\\
Climate Change	&	\href{https://www.kaggle.com/datasets/edqian/twitter-climate-change-sentiment-dataset}{Twitter Climate Change Sentiment Dataset}	&	April 2015 - Feb 2018	&	&500	\\
	&	\href{https://www.kaggle.com/datasets/die9origephit/climate-change-tweets}{Climate Change Tweets (2022)}	&	Jan - Feb 2022	& MIT	&250	\\
	&	\href{https://github.com/metunlp/MiDe22/tree/main}{MiDe22} (1 event)	&	Sept 2020	& MIT	&50	\\
	&	SemEval-2016 Task 7	&	2015 - 2016	& 	& 200	\\
Abortion	&	\href{https://data.world/crowdflower/progressive-issues-sentiment}{Progressive Issues Sentiment}	&	2015 - 2016	& CC0	&300	\\
	&	SemEval-2016 Task 7	&	2015 - 2016	& 	& 700	\\
Feminism	&	\href{https://data.world/crowdflower/progressive-issues-sentiment}{Progressive Issues Sentiment}	&	2015 - 2016	& CC0	&300	\\
	&	SemEval-2016 Task 7	&	2015 - 2016	& 	&700	\\
Refugees and Migrants	&	\href{https://github.com/metunlp/MiDe22/tree/main}{MiDe22} (9 events)	&	Dec 2020 - Jan 2022	& MIT	&540	\\
	&	Keyword Tweets 	&	Jan - Feb 2021	&	&460	\\
Brexit	&	\href{https://www.kaggle.com/datasets/visalakshiiyer/twitter-data-brexit}{Brexit Polarity Tweets} (Anti + Pro)	&	Jan - March 2022	&	CC0 &1000	\\
War in Ukraine	&	\href{https://github.com/metunlp/MiDe22/tree/main}{MiDe22} (10 events)	&	Feb - March 2022	&	MIT & 1000	\\
\bottomrule
\end{tabular}
\caption{The 11,000 human-written posts in the dataset. Sources are linked to the original dataset. Some sources require free sign-in at \url{https://data.world/}.}
\label{tab:human_data}
\end{table*}
\setlength{\tabcolsep}{6pt}

\begin{table*}[tbh]
\centering
\footnotesize
\begin{tabular}{ll}
\toprule
\textbf{Topic}  & \textbf{Collection Tags} \\
\midrule
Data Privacy	&	NSA, NSA \& spying, privacy \& leak, data \& leak, Equifax \& data breach, \\
& Cambridge Analytica	\\
Climate Change	&	\#ClimateChange, \#GlobalWarming, \#ClimateChangeScam, \#GlobalWarmingHoax, \\
& \#JunkScience, \#GlobalCooling, \#GlobalWarmingIsNotReal	\\
Abortion	&	\#Prochoice, \#Abortion, \#Prolife, \#PrayToEndAbortion, \#EndAbortion, \\
& \#PlannedParenthood	\\
Feminism	&	\#Feminism, \#FeministsAreUgly, \#INeedFeminismBecause, \#WomenAgainstFeminism, \\
& \#FeminismIsAwful	\\
Refugees and Migrants &	refugees are, migrants are	\\
\bottomrule
\end{tabular}
\caption{Known collection tags for the dataset of human-written posts.}
\label{tab:collection_tags}
\end{table*}

\begin{table*}[!t]
\centering
\footnotesize
\begin{tabular}{ll}
\toprule
\textbf{Topic}  & \textbf{Event} \\
\midrule
COVID-19	&	\href{https://www.politifact.com/factchecks/2022/feb/14/facebook-posts/covid-19-vaccines-do-not-contain-hiv/}{COVID-19 vaccines do not contain HIV.}	\\
	&	\href{https://www.usatoday.com/story/news/factcheck/2022/02/17/fact-check-covid-19-vaccines-not-linked-cancer-spike-experts-say/6711367001/}{No evidence of cancer spike linked to COVID-19 vaccines.}	\\
	&	\href{https://www.usatoday.com/story/news/factcheck/2022/02/23/fact-check-article-falsely-claims-pcr-tests-vaccinate-against-covid/6832471001/}{PCR tests diagnose COVID-19 aren't used to secretly vaccinate people.}	\\
	&	\href{https://www.politifact.com/factchecks/2022/mar/01/mary-felzkowski/no-vaccination-does-not-cause-covid-19-mutate-fast/}{Over-vaccination causes faster mutation of the (COVID-19) virus.}	\\
	&	\href{https://www.politifact.com/factchecks/2022/jan/10/tweets/social-media-posts-try-connect-sagets-death-covid-/}{Bob Saget died from the COVID-19 vaccine.}	\\
	&	\href{https://www.politifact.com/factchecks/2021/oct/21/facebook-posts/covid-19-tests-are-not-part-conspiracy-microchip-p/}{Corona PCR `test’ is implanting a microchip.}	\\
	&	\href{https://www.politifact.com/factchecks/2022/jan/05/blog-posting/old-video-clip-pfizer-ceo-not-referring-microchips/}{Pfizer CEO: New Pill Will Have a Microchip That Transmits Info Once You Swallow It!}	\\
	&	\href{https://www.politifact.com/factchecks/2021/dec/23/tweets/twitter-post-misconstrues-what-who-director-genera/}{WHO Director-General: The vaccines are being used to kill children.}	\\
	&	\href{https://www.politifact.com/factchecks/2021/dec/03/instagram-posts/covid-19-vaccines-are-not-gene-therapy-and-forbes-/}{COVID-19 vaccines are gene therapy and a recent Forbes article proves that.}	\\
	&	\href{https://www.politifact.com/factchecks/2021/nov/19/blog-posting/no-covid-19-vaccines-dont-contain-luciferase/}{COVID-19 vaccines contain luciferase.}	\\
Refugees and	&	\href{https://www.politifact.com/factchecks/2022/feb/01/facebook-posts/workers-shown-video-arriving-florida-hotel-are-her/}{Hundreds of `illegals’ were dropped off at a Florida hotel.}	\\
Migrants	&	\href{https://www.politifact.com/factchecks/2021/sep/03/ted-cruz/7000-covid-positive-migrants-quarantined-mcallen-a/}{The Biden administration has released over 7000 illegal aliens who were COVID-positive.}	\\
	&	\href{https://www.politifact.com/factchecks/2021/jul/27/facebook-posts/no-evidence-viral-claim-22-million-illegal-aliens-/}{There are 22 million illegal aliens living in America and ... voting illegally.}	\\
	&	\href{https://www.politifact.com/factchecks/2021/jul/20/terri-bryant/downstate-senator-wrong-medicaid-eligibility-illin/}{Undocumented immigrants aged over 55 may be eligible for Medicaid while citizens are not.}	\\
	&	\href{https://www.politifact.com/factchecks/2021/apr/07/blog-posting/no-report-does-not-say-biden-administration-busing/}{Biden administration busing `thousands’ of illegals to red states to get permanent status.}	\\
	&	\href{https://www.politifact.com/factchecks/2021/oct/31/mary-miller/no-migrant-caravan-tapachula-mexico-not-population/}{The migrant caravan from Tapachula Mexico ``is the population of Minneapolis.”}	\\
	&	\href{https://euvsdisinfo.eu/report/migrants-on-polish-border-were-able-to-enter-belarus-thanks-to-the-visa-free-regime}{Migrants on Polish border were able to enter Belarus thanks to the visa-free regime.}	\\
	&	\href{https://euvsdisinfo.eu/report/a-sea-of-unknown-migrants-graves-is-located-at-the-polish-border}{A sea of unknown migrants' graves is located at the Polish border.}	\\
	&	\href{https://euvsdisinfo.eu/report/turkey-introduce-isis-eu-terrorist-attacks-christmas}{Turkey may introduce ISIS members in the EU for terrorist attacks at Christmas.}	\\
Climate Change	&	\href{https://www.politifact.com/factchecks/2020/sep/15/facebook-posts/climate-change-created-conditions-west-coast-fires/}{West Coast wildfires are due to “bad forest maintenance” not climate change.}	\\
War in Ukraine	&	\href{https://www.politifact.com/factchecks/2022/mar/03/russian-embassy-canada/russias-pants-fire-claim-it-not-occupying-ukrainia/}{Russia Embassy in Canada claim that it is not occupying Ukrainian territory.}	\\
	&	\href{https://www.usatoday.com/story/news/factcheck/2022/02/21/fact-check-arma-3-not-russia-ukraine-conflict-shown-viral-video/6879521001/}{Viral clip shows `Arma 3' video game, not war between Russia and Ukraine.}	\\
	&	\href{https://www.politifact.com/factchecks/2022/feb/25/vladimir-putin/putin-repeats-long-running-claim-genocide-ukraine/}{Ethnic Russians face ``genocide perpetrated by the Kyiv regime.”}	\\
	&	\href{https://www.politifact.com/factchecks/2022/mar/04/viral-image/no-isnt-photo-russian-tank-sale-ebay/}{Photo shows a Russian tank Ukrainians are selling on eBay.}	\\
	&	\href{https://www.republicworld.com/world-news/russia-ukraine-crisis/zelensky-in-kyiv-has-not-fled-to-poland-ukrainian-parliament-counters-russias-claim-articleshow.html}{Zelensky In Kyiv Has Not Fled To Poland: Ukrainian Parliament Counters Russia's Claim.}	\\
	&	\href{https://www.politifact.com/factchecks/2022/mar/04/facebook-posts/evidence-russias-war-ukraine-plentiful-and-proves-/}{Evidence of Russia’s war in Ukraine is plentiful and proves it is not `staged’.}	\\
	&	\href{https://www.politifact.com/factchecks/2022/mar/07/facebook-posts/no-former-ukrainian-beauty-queen-has-not-joined-uk/}{Miss Universe-Ukraine turned in her high heels for combat boots to fight for her country.}	\\
	&	\href{https://www.politifact.com/factchecks/2022/mar/10/viral-image/no-isnt-real-time-cover-showing-putin-hitlers-must/}{A Time magazine cover shows Vladimir Putin with Hitler’s mustache.}	\\
	&	\href{https://www.politifact.com/factchecks/2022/mar/01/facebook-posts/samuyil-hyde-internet-meme-not-ghost-kyiv-pilot/}{Ukrainian fighter-ace known as the Ghost of Kyiv's real name is Samuyil Hyde.}	\\
	&	\href{https://www.usatoday.com/story/news/factcheck/2022/03/03/fact-check-no-evidence-putin-banned-rothschild-family-russia/6979745001/}{Vladimir Putin has banned the Rothschild family from entering Russia.}	\\
\bottomrule
\end{tabular}
\caption{MiDe22 source dataset events by Topic. Events are linked to the related fact-checking article.}
\label{tab:MiDe_events}
\end{table*}

\section{Data Generation}
\label{app:generation}

Here we list the exact prompts used to generate tweets, and provide further details on the generation model fine-tuning. Examples of output from each of the LLMs in response to each prompting strategy can be seen in Table~\ref{tab:example_outputs}.

\begin{table*}[tbhp]
    \centering
    \footnotesize
\begin{tabular}{l  p{13cm}}
\toprule
\textbf{Source} & \textbf{Output} \\
\midrule 
Human & @Noin007 though I'm not a girl so maybe I'm not  qualified to comment on women's issues, according to \#feminism\\
\midrule 
\multicolumn{2}{p{15cm}}{\textbf{Paraphrase prompt:} ``Generate the text similar to the input social media text but using different words and sentence composition. Input: <human-written tweet> Output:'' (\textit{Output after third iteration})}  \\
\midrule 
Llama-1B & Real conversations spark growth and break the glass of limitations. Join us in our quest to redefine the notion of equality. Ask the questions that will change everything. \#WomenInPower \#EqualityInAction \\
Llama-8B & @RebelHeart22 Recognizing the unique perspectives that come with non-traditional identities, I'm thoughtful in my online engagement to ensure my views are considerate of others' diverse experiences \#empatheticcommunication\\
GPT-4o & @Noin007 While I might not be the best person to talk about women's matters because I'm not a woman, that's exactly what \#feminism is all about.\\
GPT-4o-mini & @Noin007 Even though I'm not a woman, I guess I'm not really in a position to share my views on matters concerning women's rights, in line with \#feminism principles.\\

\midrule 
\multicolumn{2}{p{15cm}}{\textbf{Generate from example prompt:} ``Given the input social media text, generate 10 other posts that communicate the same information, but using different words and sentence composition. Output the 10 posts in a Python list format, with no additional text. Input: <human-written tweet> Output:'' (\textit{First text in output list})} \\
\midrule 
Llama-1B & @Noin007 Women's rights aren't just for girls\\
Llama-8B & Despite not identifying as a woman, I'm aware that I may not be the most authoritative voice on women's rights, as per some feminist ideologie \\
GPT-4o & @Noin007 While I don't identify as female, perhaps I'm not the best to speak on women's topics according to \#feminism.\\
GPT-4o-mini & @Noin007 Even though I'm not a woman, perhaps I shouldn’t weigh in on issues that affect women, based on \#feminism\\ 

\midrule
\multicolumn{2}{p{15cm}}{\textbf{Generate from topic prompt:} ``Write a tweet in casual, social media style based on the following description: The main topic is commenting on women's issues and feminism. The author expresses skepticism or criticism towards the idea that only women can comment on women's issues.''} \\
\midrule 

Llama-1B & Women's issues are just a buzzword. We've been using it to silence women's voices for too long. Why should we only be given space to say women's issues? Where are the men's voices? It's time to listen to and amplify each other's perspectives. \#Feminism \#EqualityForAll\\
Llama-8B & I'm calling BS on the notion that only women can comment on women's issues I'm not trying to take away anyone's agency, but can't we all have a seat at the table? Men, listen up - you've got a role to play in creating a more just, equal society Don't get left out, guys \#FeminismNotJustForWomen \#AlliesWanted \\
GPT-4o & Why is it that only women can talk about women’s issues? \thinkingface Feminism should be about inclusivity and open dialogue! Let’s all learn from each other and make a difference together. \biceps \sparkles \#FeminismForAll \#OpenDialogue\\
GPT-4o-mini & Why do some peeps think only women can comment on women’s issues? \thinkingface Like, we all live in this world together! Allies can have a voice too! \speakinghead Let’s lift each other up, no matter who you are! \#Feminism \#AlliesMatter \purpleheart \sparkles
\\ 

FT\_Llama-1B & @IveMaureen @Anastasia\_Lyons You're welcome. It really doesn't exist that only women can comment on women’s issues. It's open to men who have a deep understanding of what those issues are.\\
FT\_Llama-8B & @TovaCarson @tomitarell I don't see how you can say that only women can comment on women’s issues. Can men comment on human rights issues?\\
FT\_GPT-4o & @Bowie\_May Well I suppose if only women can comment on women’s issues then feminists (of which I am one) should just shut up. Historian here by the way. \\
FT\_GPT-4o-mini & Only women can comment on womens issues. Got it. \\

\bottomrule
\end{tabular}
\caption{Examples of texts produced by different models in response to the different prompting strategies.}
\label{tab:example_outputs}
\end{table*}

\subsection{Paraphrase}
 The paraphrasing prompt is taken directly from \cite{macko2024multisocial} and is as follows:

\begin{quote}
    \texttt{prompt = "Task: Generate the text similar to the input social media text but using different words and sentence composition.\\
    Input: <human-written tweet>\\ 
    Output: "}
\end{quote}

\subsection{Generate From Example}

In this strategy, we prompt the LLMs to generate 10 variations given a single human-written tweet, as follows:

\begin{quote}
    \texttt{prompt = "Task: Given the input social media text, generate 10 other posts that communicate the same information, but using different words and sentence composition. Output the 10 posts in a Python list format, with no additional text. \\
    Input: <human-written tweet> \\
    Output: "}    
\end{quote}

\subsection{Generate From Topic}
\label{app:gen_from_topic}

 We propose a new method for deriving human-AIGT topic-matched pairs that does not rely on paraphrasing. Firstly, the main topic and stance are extracted for each human-written tweet by prompting GPT-4o as follows:

\begin{quote}
    \texttt{prompt = "What is the main topic of this tweet, and what stance does the author take? Answer as concisely as possible. <human-written tweet>"}
\end{quote}

The resulting descriptions summarize the human-written tweets succinctly, for example:

\begin{quote}
    \texttt{Topic = "The main topic of the tweet is a criticism of Joe Biden and the Democrats. The author takes a stance supporting Trump's attack, suggesting that Biden and the Democrats are hypocritical in their approach to law enforcement and religious freedom."}
\end{quote}

Secondly, each LLM is instructed to write a social media post expressing the \verb|Topic| as follows:
\begin{quote}
    \texttt{prompt = "\{role: system, 
  content: You are an assistant to help write text in a casual social media style.\}, \\
  \{role: user,
  content: Write a tweet in casual, social media style based on the following description: <Topic>.\}"}
\end{quote}

\subsection{Fine-tuning}
\label{app:data_gen:fine-tuning}
We fine-tune base language models to adopt more relaxed social-media style language by observing training pairs of (\verb|Topic|, \verb|human-written tweet|). More specifically, each fine-tuning training example is instruction-formatted as:

\begin{quote}
    \texttt{train = "\{role: system, 
  content: You are an assistant to help write text in a casual social media style.\}, \\
  \{role: user,
  content: Write a tweet in casual, social media style based on the following description: <Topic>.\}, \\
  \{role: assistant, content: <human-written tweet>.\}"}
\end{quote}
where human-written tweets are sampled from an auxiliary dataset of general tweets on a random variety of topics, and \verb|Topic| is extracted from each tweet using GPT-4o as above.    

Base models GPT-4o and GPT-4o-mini were fine-tuned using OpenAI's online fine-tuning interface. This graphical user interface makes it very easy for non-technical users to create fine-tuned models. The downside is that the precise details of how the fine-tuning is accomplished is not available to the user. Furthermore, we observe a nonlinear trend in the fine-tuning time with respect to the dataset size and number of training epochs, indicating that the fine-tuning time likely depends on external (hidden) factors such as the number of other requests in the queue and the available computational resources. The other constraint in using the OpenAI interface is the cost, which is currently \$25 USD per 1 million training tokens for the GPT-4o model and \$3 USD per 1 million training tokens for GPT-4o-mini.

Base models Llama-3-8B-Instruct\footnote{\href{https://huggingface.co/meta-llama/Meta-Llama-3-8B-Instruct}{https://huggingface.co/meta-llama/Meta-Llama-3-8B-Instruct}} and Llama-3.2-1B-Instruct\footnote{\href{https://huggingface.co/meta-llama/Llama-3.2-1B-Instruct}{https://huggingface.co/meta-llama/Llama-3.2-1B-Instruct}} were fine-tuned using QLoRA (loaded in 4-bit quantization for faster training) \cite{qlora, lora}, with an adapter rank of 64 and a scaling factor of 16, for 5 epochs using a single T4 GPU\footnote{Training was conducted using a Google Colab environment (an online environment enabling easy access to remote GPUs); between Sept - Nov 2024 the average observed rate to connect to the T4 GPU runtime was 1.51 credits/hour, and the current cost is \$9.99 USD for 100 compute credits.} on two different sizes of training sets as shown in Table~\ref{tab:finetuned_models}. The smallest of the Llama-3-family models requires only 2.2 GB of GPU RAM to be fine-tuned and can be trained on a small dataset in approximately 10 minutes. This model can represent an agile generation scenario where an actor may want to train dynamically on breaking news.

\begin{table*}[!t]
\centering
\footnotesize
\begin{tabular}{llrrr}
\toprule
\textbf{Name} & \textbf{Base Model} &         \textbf{Data Size (Train)} &                \textbf{Time (minutes)} & \textbf{Cost (\$)}  \\
\midrule
FT\_Llama-8B-small	&   Llama-3-8B-Instruct	&	800	&	42	&	0.12 \\ 
FT\_Llama-8B-large	&	Llama-3-8B-Instruct	&   2000	&	108	&	0.30 \\
FT\_Llama-1B-small	&	Llama-3.2-1B-Instruct	&   800	&	10	&	0.03 \\ 
FT\_Llama-1B-large	&	Llama-3.2-1B-Instruct	&   2000	&	25	&	0.07\\ 

FT\_GPT-4o-small	&   GPT-4o	&	800	&	184	&	6.13 \\ 
FT\_GPT-4o-large	&	GPT-4o	&   2000	&	79 	& 26.11 \\ 
FT\_GPT-4o-mini-small	&	GPT-4o-mini	&   800	&	49	&	0.74 \\ 
FT\_GPT-4o-mini-large	&	GPT-4o-mini	&   2000	&	38	& 3.13\\	

\bottomrule
\end{tabular}
\caption{Fine-tuned models with training time and approximate cost (USD).}
\label{tab:finetuned_models}
\end{table*}

\subsection{Dataset Processing}
\label{app:data_gen:processing}

Firstly, we observe that the generators sometimes refuse to generate a tweet on a given topic, particularly in the case of the base models. In such cases, the generated refusal and the matching human tweet are removed from the dataset.  
Secondly, as discussed in detail in Section~\ref{app:human}, the dataset of collected human tweets contains strong identifying signals due to the specific collection criteria. For example, all human tweets on the topic of Feminism contain certain hashtags (at least one of \#Feminism, \#FeministsAreUgly, \#INeedFeminismBecause, \#WomenAgainstFeminism, \#FeminismIsAwful). Because the AI-generated text does not uniformly adhere to these same conditions, detection appears to be trivial if the full set of AIGT is included in the classification experiments. To account for this bias in the human data distribution, the same collection criteria are applied to the AIGT on a pairwise basis. That is, whenever a human tweet meets a certain collection condition, the corresponding AIGT must also meet that condition in order for \textit{both} samples from the pair to be included. The result is a balanced dataset where the distribution of collection artifacts should be similar between both classes. The final size of each dataset after filtering is summarized in Table \ref{tab:dataset_size}. Note that the ``generate 10'' (Gen-10) datasets have been condensed to one extracted generation per human example. We take the first extracted generation as a representative of this category, with no significant impact on the reported results, as discussed in Section \ref{app:AIGT_detection_results}. The total number of AI-generated tweets in the full dataset, taking all 10 extractions for every ``generate 10'' prompt, is 505,159. Note that in all classification experiments, 6000 samples from each class were taken as the training set, except in cases where the total dataset size does not permit this (Llama-3.2-1B paraphrase and generate-10 sets), where 3000 samples from each class are used instead. Re-running classification experiments on the larger datasets with either 6000 or 3000 training samples showed no significant differences. We additionally apply minimal processing to the generated tweets to remove the pre- and post-scripts that are typical of base models (e.g., ``\textit{Sure! Here is your tweet:}'' or ``\textit{Let me know if there is anything else I can help you with!}''). 

\begin{table}[t!]
    \footnotesize
\begin{tabular}{llr}
\toprule
\textbf{LLM} & \textbf{Prompt} & \textbf{Dataset Size}\\
\midrule
Llama-3.2-1B & Para-1  &  11,776  \\
Llama-3.2-1B & Para-2  &  9,298  \\
Llama-3.2-1B & Para-3  &   8,148 \\
Llama-3.2-1B & Gen-10  &   10,118  \\
Llama-3-8B & Para-1  &  14,288 \\
Llama-3-8B & Para-2  &  13,098 \\
Llama-3-8B & Para-3  &  12,356 \\
Llama-3-8B & Gen-10  &  17,338 \\
GPT-4o-mini & Para-1  &  18,070 \\
GPT-4o-mini & Para-2  &  17,502 \\
GPT-4o-mini & Para-3  &  16,874 \\
GPT-4o-mini & Gen-10  & 20,574 \\
GPT-4o & Para-1  &  17,298 \\
GPT-4o & Para-2  &  16,626 \\
GPT-4o & Para-3  &  15,912 \\
GPT-4o & Gen-10  &  20,560 \\
\hline
\midrule
Llama-3.2-1B & Topic  &  14,406 \\
FT\_Llama-1B-small & Topic  & 12,898  \\
FT\_Llama-1B-large & Topic  & 13,546 \\
\midrule
Llama-3-8B & Topic  & 16,296 \\
FT\_Llama-8B-small & Topic  & 14,212 \\
FT\_Llama-8B-large & Topic  & 13,914 \\
\midrule
GPT4o-mini & Topic  & 16,884 \\
FT\_GPT4o-mini-small & Topic  & 14,400 \\
FT\_GPT4o-mini-large & Topic  & 14,422 \\
\midrule
GPT4o & Topic  & 16,460\\
FT\_GPT4o-small & Topic  & 14,702\\
FT\_GPT4o-large & Topic  &  14,380 \\
\bottomrule
\end{tabular}
\caption{The total size of all datasets after filtering is applied to account for generator refusal and bias in the human data distribution. All datasets are balanced between the human-written and AI-generated classes.}
\label{tab:dataset_size}
\end{table}

Furthermore, all tweets in the dataset (both human-written and AI-generated) have been post-processed by removing all mentions (@'s) and links. We observed that the fine-tuned generators were not successful at producing realistic mentions and links that matched the human-written distribution, and therefore the inclusion of these features artificially made the AIGT more detectable. Here we are interested in studying the linguistic ability of the generating models in producing social-media style language (including words, hashtags, and emojis). We assume that any reasonably sophisticated actor could insert the desired mentions and relevant links as a separate step, potentially using a different purpose-built model for this subtask. For reference, the effect of this post-processing step on the detectability of the generated tweets is shown in Table \ref{tab:detectsbility_data_condition}.

\section{Linguistic Analysis}
\label{app:linguistic_analysis}

\begin{figure*}[htbp]
    \centering
    \begin{subfigure}[b]{0.23\textwidth}
        \centering
        \includegraphics[width=\textwidth]{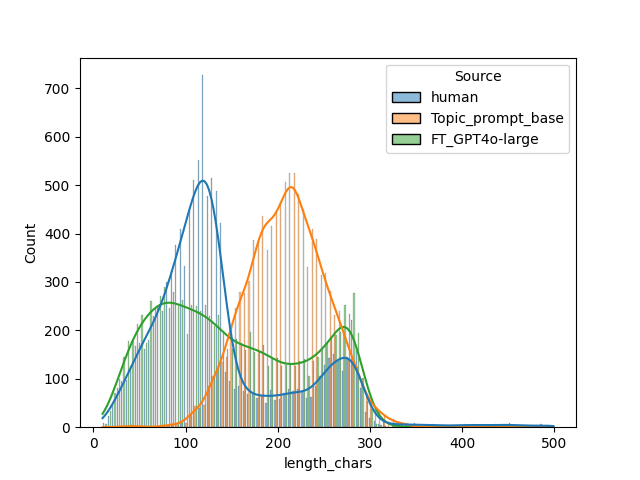}
        \caption{GPT-4o}
    \end{subfigure}
    \begin{subfigure}[b]{0.23\textwidth}
        \centering
        \includegraphics[width=\textwidth]{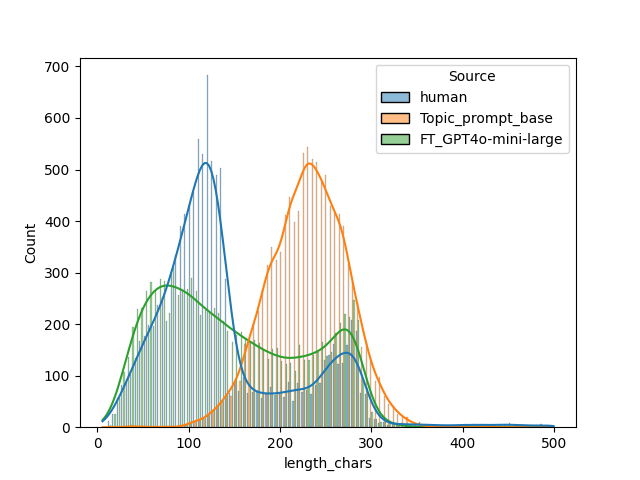}
        \caption{GPT-4o-mini}
    \end{subfigure}
    \begin{subfigure}[b]{0.23\textwidth}
        \centering
        \includegraphics[width=\textwidth]{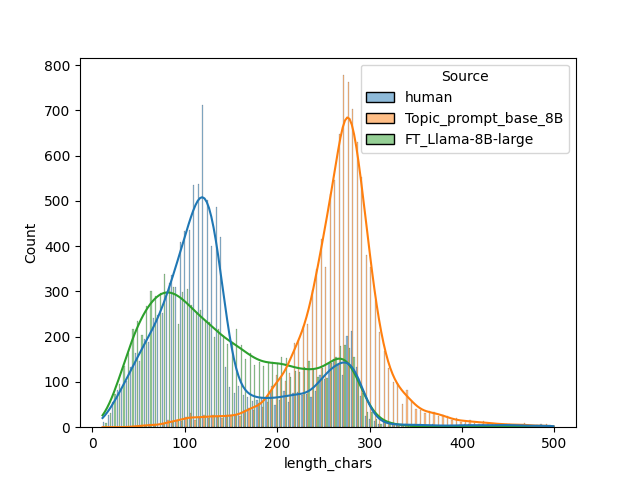}
        \caption{Llama-3-8B}
    \end{subfigure}
    \begin{subfigure}[b]{0.23\textwidth}
        \centering
        \includegraphics[width=\textwidth]{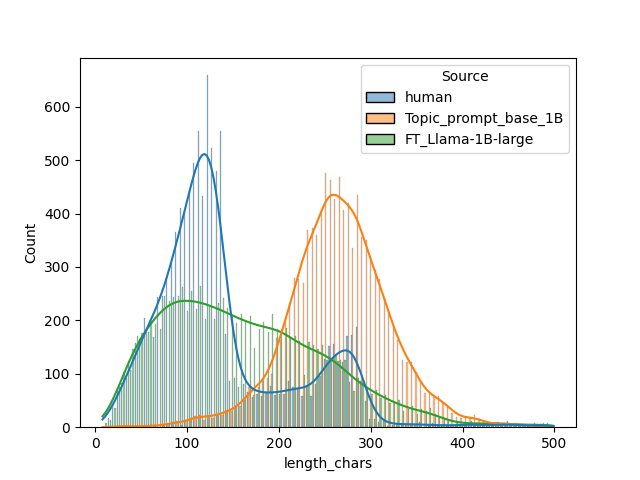}
        \caption{Llama-3.2-1B}
    \end{subfigure}

    \vspace{0.5cm}  

    \begin{subfigure}[b]{0.23\textwidth}
        \centering
        \includegraphics[width=\textwidth]{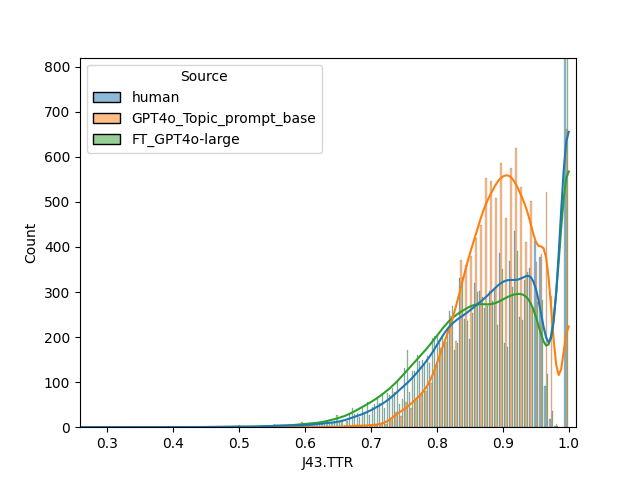}
        \caption{GPT-4o}
    \end{subfigure}
    \begin{subfigure}[b]{0.23\textwidth}
        \centering
        \includegraphics[width=\textwidth]{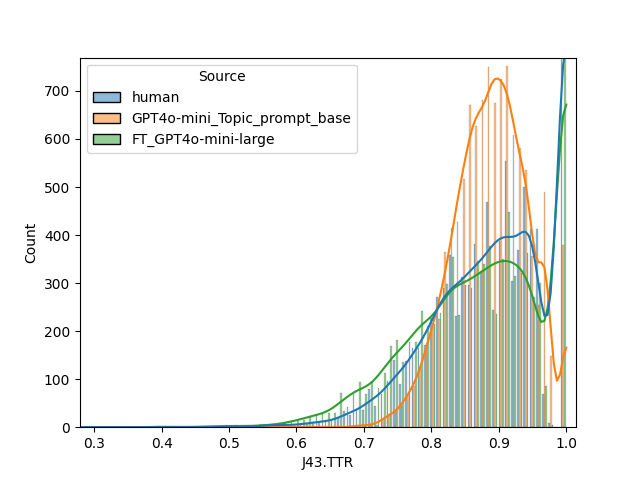}
        \caption{GPT-4o-mini}
    \end{subfigure}
    \begin{subfigure}[b]{0.23\textwidth}
        \centering
        \includegraphics[width=\textwidth]{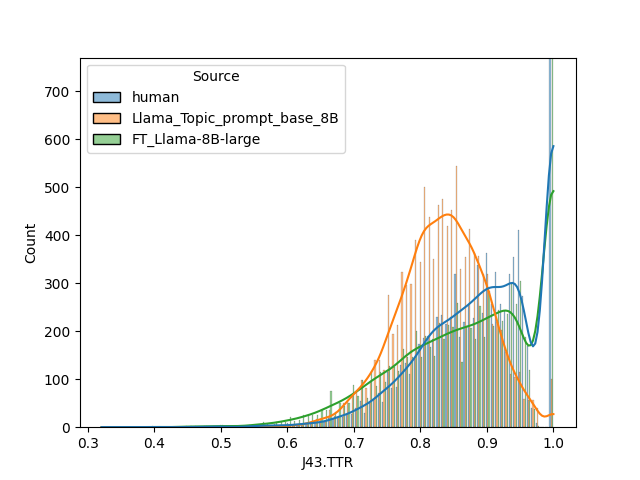}
        \caption{Llama-3-8B}
    \end{subfigure}
    \begin{subfigure}[b]{0.23\textwidth}
        \centering
        \includegraphics[width=\textwidth]{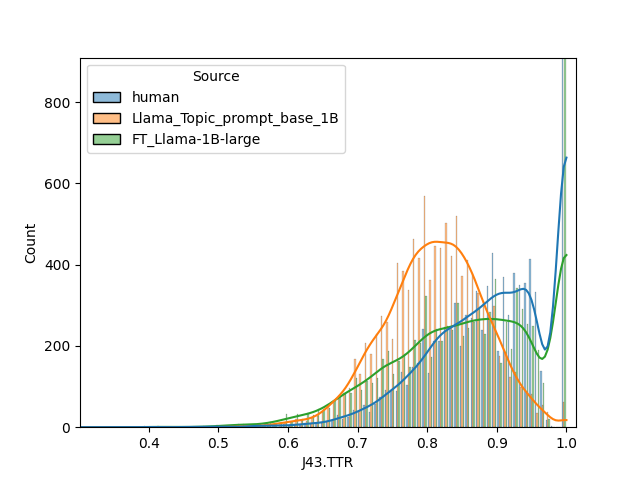}
        \caption{Llama-3.2-1B}
    \end{subfigure}

    \vspace{0.5cm}  

    \begin{subfigure}[b]{0.23\textwidth}
        \centering
        \includegraphics[width=\textwidth]{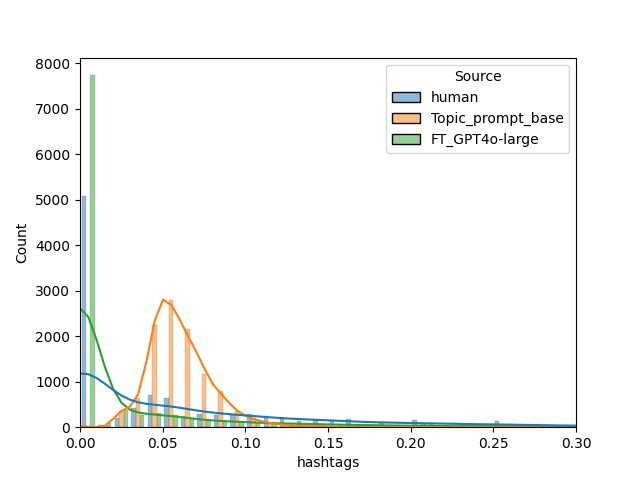}
        \caption{GPT-4o}
    \end{subfigure}
    \begin{subfigure}[b]{0.23\textwidth}
        \centering
        \includegraphics[width=\textwidth]{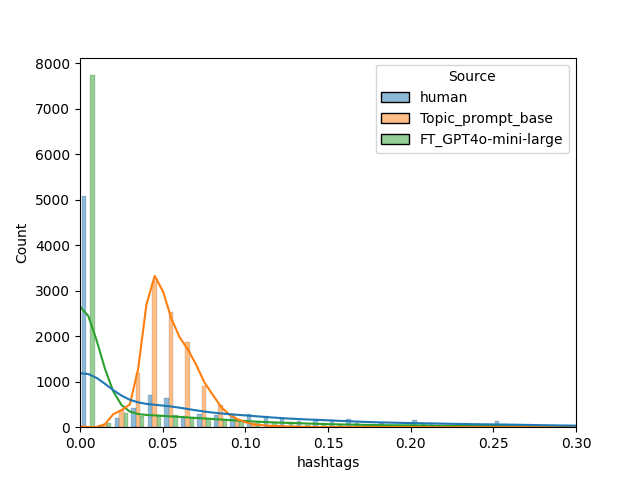}
        \caption{GPT-4o-mini}
    \end{subfigure}
    \begin{subfigure}[b]{0.23\textwidth}
        \centering
        \includegraphics[width=\textwidth]{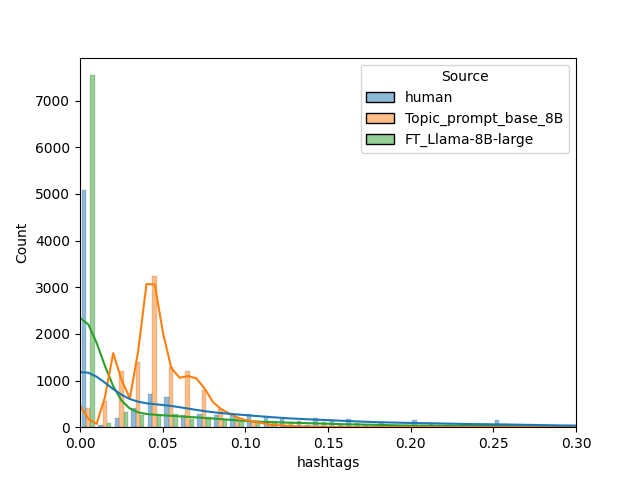}
        \caption{Llama-3-8B}
    \end{subfigure}
    \begin{subfigure}[b]{0.23\textwidth}
        \centering
        \includegraphics[width=\textwidth]{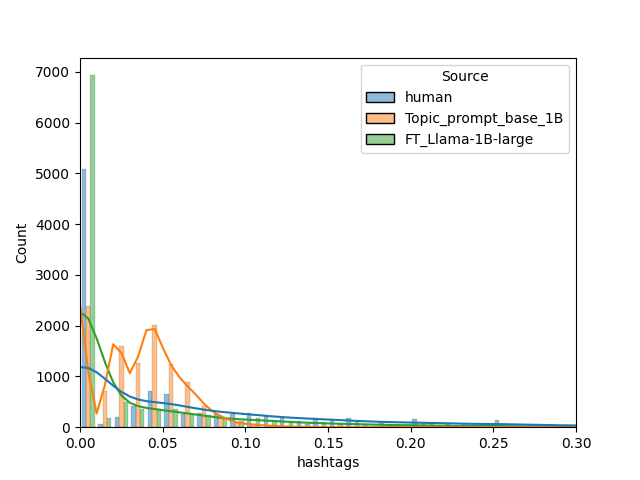}
        \caption{Llama-3.2-1B}
    \end{subfigure}

    \caption{Length of text in characters (top), Type-Token Ratio or TTR (middle), and number of hashtags (bottom), three features which vary significantly between the base models and the human-written data but become closer after fine-tuning.}
    \label{fig:linguistic_dist}
\end{figure*}

Here we provide additional details on the linguistic analysis. Table~\ref{tab:effect_size} summarized the non-zero effect sizes for GPT-4o only; the full results for all four models are available in Tables~\ref{tab:effect_size_gpt4o}, \ref{tab:effect_size_gpt4o-mini}, \ref{tab:effect_size_llama-8b}, \ref{tab:effect_size_llama-1b}. More information about Biber's set of linguistic features can be found in  \citet{biber-1993-using} and the Appendix of \citet{reinhart2024llms}. Features were extracted using the BiberPy Python library \url{https://github.com/ssharoff/biberpy}, version downloaded on 1 December 2024, available under a GPL-3.0 license.

We identified that features which seemed to be associated with AIGT across all four models included adverbs, first-person pronouns, private verbs (verbs expressing internal state, e.g. think, believe), and contractions. Interestingly, in many cases the trends were the opposite to what we might have expected: LLMs produced \textit{more} first-person pronouns, private verbs, and contractions -- in contrast to the information-dense, formal style observed in previous studies \cite{herbold2023large, reinhart2024llms, sardinha2024ai}. One possibility that we considered is that the automated linguistic feature extraction tools, which were not trained on social media text, may be introducing errors into this measurement, unrelated to the provenance of the text. Further analysis suggests that this may be the case for contractions -- human users are more likely to omit the apostrophe from a contraction (e.g., \textit{I cant believe it!}), and this is not identified as a contraction by the POS-tagger. 

\begin{figure*}[tbh]
    \centering
    \includegraphics[width=\linewidth]{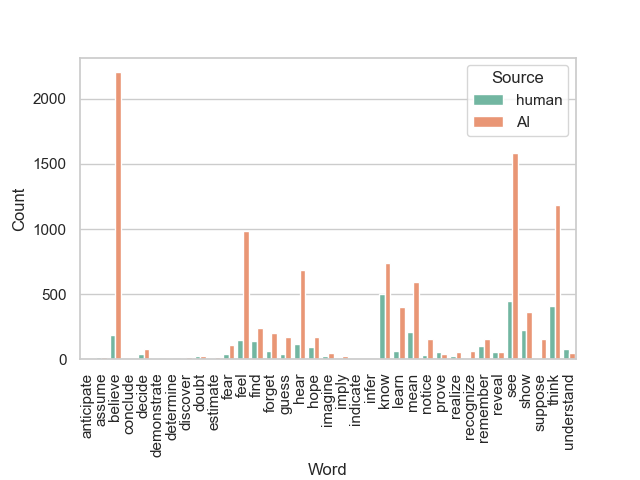}
    \caption{Frequency of occurrence of the most frequent ``private verbs'' by Llama-3-8B and humans.}
    \label{fig:private_verbs}
\end{figure*}

However, adverbs, first-person pronouns, and private verbs do appear to occur more frequently in AI-generated texts -- albeit, driven by a tendency towards particular word usages. As one example, Figure~\ref{fig:private_verbs} shows that the trend for Llama-3-8B (base model) to use more private verbs than humans is driven by an exceptionally high usage of a few particular verbs, such as \textit{believe}. Examining the texts, we observe that humans tend to use the word \textit{believe} in a variety of different contexts, while the LLM over-uses a particular phrase:

\begin{displayquote}
\textbf{Can't believe} @POTUS \& @VP are still asleep at the wheel on immigration! After 2 yrs, nothing but empty promises \& rising border crisis Our country's security \& economy suffer bc of inaction Time for IMPEACHMENT! \#ImpeachBidenHarris \#ImmigrationCrisis \#DoSomething \\
\textbf{Can't believe} the drama over the latest'migrant caravan'. Meanwhile, my aunt is living in a van down by the river and nobody's losing their minds. \#HypedForNoReason \#FakeCrisis \\
\textbf{Can't believe} @JoeBiden \& @KamalaHarris are sitting back while a massive migrant caravan makes its way to our border! Dereliction of duty, anyone? I'm done waiting for them to do their jobs. Time to hold them accountable! \#FireBiden \#HarrisMustGo \\
\textbf{Can't believe} some folks still trying to cross into Poland, despite Belarus saying `come on in, stay for free' What's the problem, folks? Minsk's got couches, not cardboard boxes! If conditions are tough, pack up and head back to Minsk, don't risk it \#PolandHasBorders
\end{displayquote}

\begin{figure*}[tbhp]
    \centering
    \includegraphics[width=\linewidth]{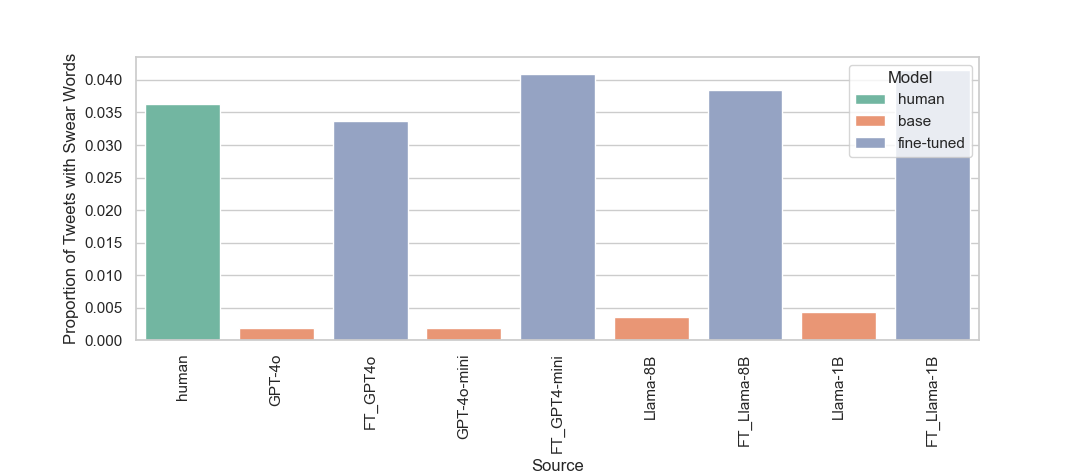}
    \caption{Proportion of tweets containing at least one profanity.}
    \label{fig:swear_words}
\end{figure*}

We also explored features that did not occur frequently enough to be included in the statistical analysis, but that nonetheless showed interesting correlations with one class or the other. One example of this is swear words (as defined by Wikipedia\footnote{\url{https://en.wikipedia.org/wiki/Category:English_profanity}}). The LLMs have been trained to (mostly) not produce profanity, and so when it occurs it almost always indicates a human text, although the overall rate of swearing in the human data is also low (around 3.5\%). However, we observed that the fine-tuned models re-learned the ability to swear (see Figure~\ref{fig:swear_words}), and at a rate remarkably similar to that in our human data, even though the models were fine-tuned on a completely different sample of Twitter data, and specifically fine-tuned on a sample that was benign enough to pass OpenAI's content moderation filters for training data. 

\begin{figure*}[tbhp]
    \centering
    \includegraphics[width=\linewidth]{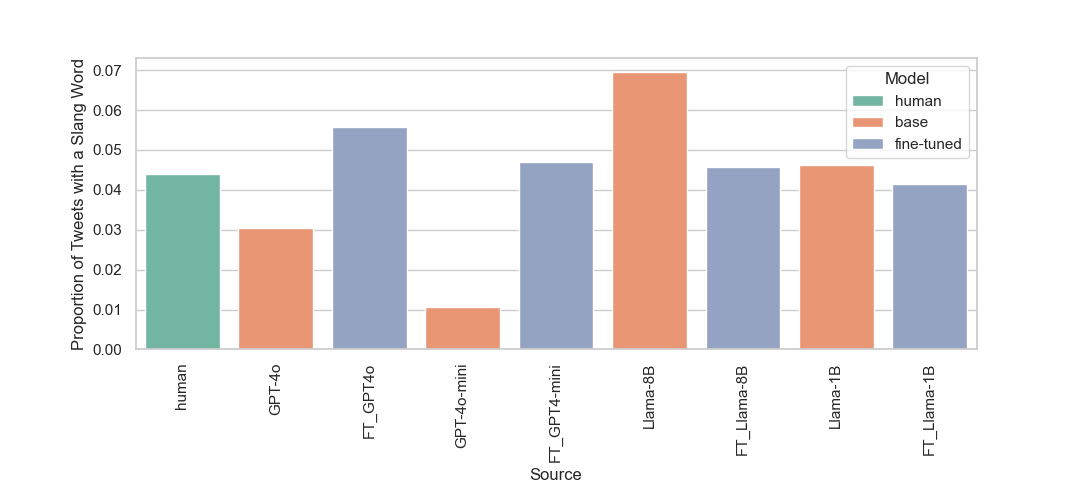}
    \caption{Proportion of tweets containing at least one slang word or abbreviation.}
    \label{fig:slang_words}
\end{figure*}

Another example of a ``rare'' linguistic feature is the use of internet slang (modified from Wikipedia\footnote{\url{https://en.wiktionary.org/wiki/Appendix:English_internet_slang}}). We had originally hypothesized that LLMs would not use as many slang words or abbreviations (e.g., \textit{wtf}, \textit{lmao}, \textit{irl}) as humans. This was true for GPT-4o and GPT-4o-mini, but not for the Llama models, which in fact produced \textit{more} slang than humans (see Fig.~\ref{fig:slang_words}). Furthermore, additional analysis revealed lexical differences between the human text and Llama-3-8B text, with Llama producing far more instances of \textit{ppl} and \textit{omg}, and humans producing instances of \textit{lol} and \textit{u} (for \textit{you}). After fine-tuning, the lexical differences are diminished.

While the focus of this paper was not to analyze linguistic properties of AI-generated text, we believe this kind of analysis provides additional insight into the impact of fine-tuning on detectability. We would emphasize that the language of social media changes rapidly, as new hashtags are introduced in response to current events, new words are devised to avoid content filters, and new names and topics of interest emerge. Therefore, an analysis of current-day social media data might uncover even larger differences between human-written text and LLM generations, due to the latter's training data time horizons.

\begin{table*}
    \footnotesize
\begin{tabular}{lrrrrr}
\toprule
Feature & Human (mean) & Base LLM (mean) & Fine-tuned (mean) & Human-Base (R) & Human-FT (R) \\
\midrule
ats & 0.043 & 0.00067 & 0.053 & \cellcolor{orange!25}-0.47 & \cellcolor{yellow!25}0.11 \\
links & 0.026 & 2.2e-05 & 0.02 & \cellcolor{orange!25}-0.38 & \cellcolor{green!25} -0.068 \\
hashtags & 0.071 & 0.06 & 0.029 & \cellcolor{yellow!25}0.26 & \cellcolor{yellow!25}-0.29 \\
emojis & 0.0074 & 0.071 & 0.011 & \cellcolor{red!25}0.92 & \cellcolor{green!25}0.048 \\
length\_chars & 1.4e+02 & 2.1e+02 & 1.5e+02 & \cellcolor{red!25}0.6 & \cellcolor{green!25}0.084 \\
offensive & 0.021 & 0.0086 & 0.021 &\cellcolor{yellow!25} -0.11 & \cellcolor{green!25}0.017 \\
misspelled & 0.017 & 0.0084 & 0.019 & \cellcolor{green!25}-0.053 & \cellcolor{green!25}0.04 \\
upper\_lower\_ratio & 0.1 & 0.066 & 0.084 & \cellcolor{yellow!25}-0.15 &\cellcolor{yellow!25} -0.18 \\

\hline 

A01.pastVerbs & 0.17 & 0.11 & 0.17 & \cellcolor{yellow!25}-0.1 & \cellcolor{green!25}0.021 \\
A03.presVerbs & 0.26 & 0.36 & 0.28 & \cellcolor{yellow!25}0.24 & \cellcolor{green!25}0.051 \\
B05.timeAdverbials & 0.0054 & 0.0041 & 0.0047 & \cellcolor{green!25}0.023 & \cellcolor{green!25}0.0036 \\
C06.1persProns & 0.025 & 0.029 & 0.022 &\cellcolor{yellow!25} 0.21 & \cellcolor{green!25}-0.025 \\
C07.2persProns & 0.012 & 0.0073 & 0.015 & \cellcolor{green!25}-0.014 & \cellcolor{green!25}0.071 \\
C08.3persProns & 0.013 & 0.0089 & 0.015 &\cellcolor{green!25} -0.0097 & \cellcolor{green!25}0.054 \\
C09.impersProns & 0.0082 & 0.015 & 0.0086 & \cellcolor{yellow!25}0.24 & \cellcolor{green!25}0.022 \\
C10.demonstrProns & 0.012 & 0.012 & 0.015 & \cellcolor{green!25}0.085 & \cellcolor{green!25}0.069 \\
C11.indefProns & 0.003 & 0.0062 & 0.0039 &\cellcolor{yellow!25} 0.14 & \cellcolor{green!25}0.035 \\
C12.doAsProVerb & 0.0034 & 0.0027 & 0.0042 & \cellcolor{green!25}0.016 & \cellcolor{green!25}0.031 \\
D13.whQuestions & 0.013 & 0.015 & 0.013 & \cellcolor{yellow!25}0.14 & \cellcolor{green!25}0.017 \\
E14.nominalizations & 0.014 & 0.015 & 0.016 & \cellcolor{yellow!25}0.11 & \cellcolor{green!25}0.053 \\
E16.Nouns & 0.14 & 0.14 & 0.14 & \cellcolor{green!25}0.068 & \cellcolor{green!25}0.04 \\
G19.beAsMain & 0.14 & 0.11 & 0.14 & \cellcolor{green!25}0.066 &\cellcolor{green!25} 0.014 \\
H23.WHclauses & 0.0098 & 0.0099 & 0.0098 & \cellcolor{green!25}0.079 & \cellcolor{green!25}0.021 \\
I39.preposn & 0.08 & 0.079 & 0.082 & \cellcolor{green!25}0.0059 & \cellcolor{green!25}0.025 \\
I40.attrAdj & 0.058 & 0.065 & 0.06 & \cellcolor{yellow!25}0.15 & \cellcolor{green!25}0.029 \\
I42.ADV & 0.036 & 0.05 & 0.039 & \cellcolor{yellow!25}0.26 & \cellcolor{green!25}0.05 \\
J43.TTR & 0.9 & 0.9 & 0.89 & \cellcolor{green!25}-0.084 & \cellcolor{green!25}-0.062 \\
J44.wordLength & 4 & 4.1 & 4.1 & \cellcolor{green!25}0.089 & \cellcolor{green!25}0.097 \\
K45.conjuncts & 0.076 & 0.069 & 0.075 & \cellcolor{green!25}0.015 & \cellcolor{green!25}0.0034 \\
K55.publicVerbs & 0.033 & 0.021 & 0.029 & \cellcolor{green!25}-0.017 & \cellcolor{green!25}-0.008 \\
K56.privateVerbs & 0.064 & 0.11 & 0.074 & \cellcolor{yellow!25}0.25 & \cellcolor{green!25}0.056 \\
L52.possibModals & 0.031 & 0.047 & 0.027 & \cellcolor{yellow!25}0.13 & \cellcolor{green!25}-0.00024 \\
L54.predicModals & 0.032 & 0.0072 & 0.029 & \cellcolor{green!25}-0.077 & \cellcolor{green!25}0.00025 \\
N59.contractions & 0.12 & 0.24 & 0.11 & \cellcolor{orange!25} 0.39 & \cellcolor{green!25}-0.02 \\
N60.thatDeletion & 0.087 & 0.12 & 0.092 & \cellcolor{yellow!25}0.18 & \cellcolor{green!25}0.034 \\
P67.analNegn & 0.099 & 0.09 & 0.11 & \cellcolor{green!25}0.074 & \cellcolor{green!25}0.046 \\

\bottomrule
\end{tabular}
    
\caption{Means and Effect size (R) of differences between human-written texts and texts generated by \textbf{GPT-4o} (base model and fine-tuned (FT)). Positive R values indicate that the feature value is higher in AIGT than in human-written text. Effect size can be interpreted as: $|R| < 0.1$ - \colorbox{green!25}{no effect}, $|R| > 0.1$ - \colorbox{yellow!25}{small effect}, $|R| > 0.3$ - \colorbox{orange!25}{medium effect}, $|R| > 0.5$ - \colorbox{red!25}{large effect}.}
\label{tab:effect_size_gpt4o}
\end{table*}

\begin{table*}
    \footnotesize
\begin{tabular}{lrrrrr}
\toprule
Feature & Human (mean) & Base LLM (mean) & Fine-tuned (mean) & Human-Base (R) & Human-FT (R) \\
\midrule
ats & 0.042 & 0.00081 & 0.049 & \cellcolor{orange!25}-0.46 & \cellcolor{yellow!25}0.11 \\
links & 0.026 & 0 & 0.024 & \cellcolor{orange!25}-0.38 & \cellcolor{green!25}-0.021 \\
hashtags & 0.071 & 0.054 & 0.03 & \cellcolor{yellow!25}0.22 & \cellcolor{yellow!25}-0.28 \\
emojis & 0.0074 & 0.083 & 0.012 & \cellcolor{red!25}0.94 & \cellcolor{green!25}0.047 \\
length\_chars & 1.4e+02 & 2.3e+02 & 1.5e+02 & \cellcolor{red!25}0.68 & \cellcolor{green!25}0.057 \\
offensive & 0.021 & 0.0085 & 0.022 & \cellcolor{green!25}-0.099 & \cellcolor{green!25}0.014 \\
misspelled & 0.017 & 0.0063 & 0.016 & \cellcolor{green!25}-0.089 & \cellcolor{green!25}-0.0021 \\
upper\_lower\_ratio & 0.1 & 0.062 & 0.086 &\cellcolor{yellow!25} -0.21 & \cellcolor{yellow!25}-0.15 \\

\hline 

A01.pastVerbs & 0.17 & 0.1 & 0.16 & \cellcolor{green!25}-0.097 & \cellcolor{green!25}0.0047 \\
A03.presVerbs & 0.26 & 0.37 & 0.29 & \cellcolor{yellow!25}0.28 & \cellcolor{green!25}0.069 \\
B05.timeAdverbials & 0.0054 & 0.003 & 0.0052 & \cellcolor{green!25}-0.0046 & \cellcolor{green!25}0.0091 \\
C06.1persProns & 0.025 & 0.032 & 0.023 & \cellcolor{yellow!25}0.27 &\cellcolor{green!25} -0.011 \\
C07.2persProns & 0.012 & 0.0061 & 0.017 & \cellcolor{green!25}-0.038 & \cellcolor{green!25}0.075 \\
C08.3persProns & 0.013 & 0.0095 & 0.014 & \cellcolor{green!25}0.02 & \cellcolor{green!25}0.021 \\
C09.impersProns & 0.0082 & 0.017 & 0.0094 & \cellcolor{orange!25}0.33 & \cellcolor{green!25}0.036 \\
C10.demonstrProns & 0.012 & 0.013 & 0.015 & \cellcolor{yellow!25}0.13 &\cellcolor{green!25} 0.073 \\
C11.indefProns & 0.003 & 0.0057 & 0.0037 & \cellcolor{yellow!25}0.14 & \cellcolor{green!25}0.029 \\
D13.whQuestions & 0.013 & 0.015 & 0.013 & \cellcolor{yellow!25}0.17 &\cellcolor{green!25} 0.022 \\
E14.nominalizations & 0.014 & 0.016 & 0.016 & \cellcolor{yellow!25}0.17 & \cellcolor{green!25}0.032 \\
E16.Nouns & 0.14 & 0.13 & 0.15 & \cellcolor{green!25}-0.024 & \cellcolor{green!25}0.052 \\
G19.beAsMain & 0.14 & 0.083 & 0.13 & \cellcolor{green!25}0.0023 & \cellcolor{green!25}-0.0047 \\
H23.WHclauses & 0.0098 & 0.0094 & 0.011 & \cellcolor{green!25}0.087 & \cellcolor{green!25}0.041 \\
I39.preposn & 0.08 & 0.079 & 0.078 & \cellcolor{green!25}0.017 & \cellcolor{green!25}-0.016 \\
I40.attrAdj & 0.058 & 0.063 & 0.057 & \cellcolor{yellow!25}0.14 & \cellcolor{green!25}0.0056 \\
I42.ADV & 0.036 & 0.054 & 0.039 & \cellcolor{orange!25}0.34 & \cellcolor{green!25}0.056 \\
J43.TTR & 0.9 & 0.89 & 0.88 & \cellcolor{yellow!25}-0.15 & \cellcolor{green!25}-0.076 \\
J44.wordLength & 4 & 4.1 & 4.1 & \cellcolor{green!25}0.056 & \cellcolor{green!25}0.045 \\
K45.conjuncts & 0.076 & 0.054 & 0.075 & \cellcolor{green!25}-0.097 & \cellcolor{green!25}-0.0018 \\
K49.generalEmphatics & 0.0019 & 0.0036 & 0.0028 & \cellcolor{green!25}0.089 & \cellcolor{green!25}0.028 \\
K55.publicVerbs & 0.033 & 0.018 & 0.025 & \cellcolor{green!25}-0.023 & \cellcolor{green!25}-0.026 \\
K56.privateVerbs & 0.064 & 0.11 & 0.075 & \cellcolor{yellow!25}0.29 & \cellcolor{green!25}0.059 \\
L52.possibModals & 0.031 & 0.06 & 0.037 & \cellcolor{yellow!25}0.23 & \cellcolor{green!25}0.033 \\
L54.predicModals & 0.032 & 0.0044 & 0.028 & \cellcolor{green!25}-0.088 &\cellcolor{green!25} -0.0015 \\
N59.contractions & 0.12 & 0.097 & 0.1 & \cellcolor{green!25}0.079 & \cellcolor{green!25}-0.046 \\
N60.thatDeletion & 0.087 & 0.11 & 0.09 & \cellcolor{yellow!25}0.19 & \cellcolor{green!25}0.028 \\
P67.analNegn & 0.099 & 0.076 & 0.12 & \cellcolor{green!25}0.062 & \cellcolor{green!25}0.067 \\
\bottomrule
\end{tabular}

\caption{Means and Effect size (R) of differences between human-written texts and texts generated by \textbf{GPT-4o-mini} (base model and fine-tuned (FT)). Positive R values indicate that the feature value is higher in AIGT than in human-written text. Effect size can be interpreted as: $|R| < 0.1$ - \colorbox{green!25}{no effect}, $|R| > 0.1$ - \colorbox{yellow!25}{small effect}, $|R| > 0.3$ - \colorbox{orange!25}{medium effect}, $|R| > 0.5$ - \colorbox{red!25}{large effect}.}
\label{tab:effect_size_gpt4o-mini}
\end{table*}

\begin{table*}
    \footnotesize

\begin{tabular}{lrrrrr}
\toprule
Feature & Human (mean) & Base LLM (mean) & Fine-tuned (mean) & Human-Base (R) & Human-FT (R) \\
\midrule

ats & 0.042 & 0.0043 & 0.066 & \cellcolor{orange!25}-0.38 & \cellcolor{yellow!25}0.17 \\
links & 0.026 & 1.4e-05 & 0.023 & \cellcolor{orange!25}-0.38 & \cellcolor{green!25}-0.027 \\
hashtags & 0.071 & 0.047 & 0.034 & \cellcolor{yellow!25}0.14 & \cellcolor{yellow!25}-0.26 \\
length\_chars & 1.4e+02 & 2.7e+02 & 1.4e+02 & \cellcolor{red!25}0.79 & \cellcolor{green!25}0.02 \\
offensive & 0.021 & 0.011 & 0.02 & \cellcolor{green!25}-0.033 & \cellcolor{green!25}-0.0015 \\
misspelled & 0.017 & 0.0094 & 0.015 & \cellcolor{green!25}0.0017 & \cellcolor{green!25}-0.022 \\
upper\_lower\_ratio & 0.1 & 0.062 & 0.08 &\cellcolor{yellow!25} -0.24 & \cellcolor{yellow!25}-0.15 \\

\hline 

A01.pastVerbs & 0.17 & 0.17 & 0.16 & \cellcolor{green!25}0.091 &\cellcolor{green!25} -0.02 \\
A03.presVerbs & 0.26 & 0.33 & 0.3 & \cellcolor{yellow!25}0.18 & \cellcolor{green!25}0.097 \\
B04.placeAdverbials & 0.0018 & 0.0022 & 0.0015 & \cellcolor{green!25}0.052 & \cellcolor{green!25}-0.0062 \\
B05.timeAdverbials & 0.0054 & 0.0037 & 0.0054 & \cellcolor{green!25}0.038 & \cellcolor{green!25}0.0098 \\
C06.1persProns & 0.025 & 0.039 & 0.028 & \cellcolor{orange!25}0.37 & \cellcolor{green!25}0.04 \\
C07.2persProns & 0.012 & 0.0071 & 0.016 & \cellcolor{green!25}-0.00092 & \cellcolor{green!25}0.076 \\
C08.3persProns & 0.013 & 0.012 & 0.013 & \cellcolor{green!25}0.093 & \cellcolor{green!25}0.0017 \\
C09.impersProns & 0.0082 & 0.011 & 0.011 & \cellcolor{yellow!25}0.22 & \cellcolor{green!25}0.064 \\
C10.demonstrProns & 0.012 & 0.012 & 0.014 & \cellcolor{yellow!25}0.12 & \cellcolor{green!25}0.049 \\
C11.indefProns & 0.003 & 0.0047 & 0.0034 & \cellcolor{yellow!25}0.12 & \cellcolor{green!25}0.023 \\
C12.doAsProVerb & 0.0034 & 0.0023 & 0.0043 & \cellcolor{green!25}0.025 &\cellcolor{green!25} 0.03 \\
D13.whQuestions & 0.013 & 0.013 & 0.013 &\cellcolor{yellow!25} 0.14 & \cellcolor{green!25}0.022 \\
E14.nominalizations & 0.014 & 0.02 & 0.017 &\cellcolor{yellow!25} 0.27 & \cellcolor{green!25}0.051 \\
E16.Nouns & 0.14 & 0.15 & 0.14 & \cellcolor{yellow!25}0.15 & \cellcolor{green!25}0.00013 \\
G19.beAsMain & 0.14 & 0.13 & 0.14 & \cellcolor{yellow!25}0.16 & \cellcolor{green!25}0.0063 \\
H23.WHclauses & 0.0098 & 0.011 & 0.01 & \cellcolor{yellow!25}0.15 & \cellcolor{green!25}0.023 \\
I39.preposn & 0.08 & 0.081 & 0.078 & \cellcolor{green!25}0.033 & \cellcolor{green!25}-0.024 \\
I40.attrAdj & 0.058 & 0.066 & 0.056 &\cellcolor{yellow!25} 0.17 & \cellcolor{green!25}-0.0017 \\
I42.ADV & 0.036 & 0.047 & 0.037 & \cellcolor{yellow!25}0.27 & \cellcolor{green!25}0.02 \\
J43.TTR & 0.9 & 0.83 & 0.88 & \cellcolor{orange!25}-0.47 & \cellcolor{green!25}-0.073 \\
J44.wordLength & 4 & 4.2 & 4 &\cellcolor{yellow!25} 0.18 & \cellcolor{green!25}-0.0056 \\
K45.conjuncts & 0.076 & 0.092 & 0.078 & \cellcolor{yellow!25}0.21 & \cellcolor{green!25}0.021 \\
K55.publicVerbs & 0.033 & 0.029 & 0.025 & \cellcolor{green!25}0.04 & \cellcolor{green!25}-0.028 \\
K56.privateVerbs & 0.064 & 0.12 & 0.081 & \cellcolor{orange!25}0.34 & \cellcolor{green!25}0.072 \\
L52.possibModals & 0.032 & 0.074 & 0.034 & \cellcolor{yellow!25}0.28 & \cellcolor{green!25}0.019 \\
L54.predicModals & 0.031 & 0.015 & 0.031 & \cellcolor{green!25}-0.027 & \cellcolor{green!25}-0.003 \\
N59.contractions & 0.12 & 0.4 & 0.11 & \cellcolor{red!25}0.65 & \cellcolor{green!25}-0.037 \\
N60.thatDeletion & 0.087 & 0.14 & 0.096 & \cellcolor{orange!25}0.3 & \cellcolor{green!25}0.042 \\
P67.analNegn & 0.099 & 0.14 & 0.12 & \cellcolor{yellow!25}0.27 & \cellcolor{green!25}0.06 \\
\bottomrule
\end{tabular}

\caption{Means and Effect size (R) of differences between human-written texts and texts generated by \textbf{Llama-3-8B} (base model and fine-tuned (FT)). Positive R values indicate that the feature value is higher in AIGT than in human-written text. Effect size can be interpreted as: $|R| < 0.1$ - \colorbox{green!25}{no effect}, $|R| > 0.1$ - \colorbox{yellow!25}{small effect}, $|R| > 0.3$ - \colorbox{orange!25}{medium effect}, $|R| > 0.5$ - \colorbox{red!25}{large effect}.}
\label{tab:effect_size_llama-8b}
\end{table*}

\begin{table*}
    \footnotesize

\begin{tabular}{lrrrrr}
\toprule
Feature & Human (mean) & Base LLM (mean) & Fine-tuned (mean) & Human-Base (R) & Human-FT (R) \\
\midrule
ats & 0.042 & 0.003 & 0.066 & \cellcolor{orange!25}-0.42 & \cellcolor{yellow!25}0.15 \\
links & 0.026 & 4.8e-05 & 0.023 & \cellcolor{orange!25}-0.38 & \cellcolor{green!25}-0.018 \\
hashtags & 0.071 & 0.035 & 0.035 & \cellcolor{green!25}-0.015 & \cellcolor{yellow!25}-0.22 \\
length\_chars & 1.4e+02 & 2.7e+02 & 1.6e+02 & \cellcolor{red!25}0.78 & \cellcolor{yellow!25}0.16 \\
offensive & 0.021 & 0.011 & 0.019 & \cellcolor{green!25}-0.019 & \cellcolor{green!25}0.0075 \\
misspelled & 0.017 & 0.0085 & 0.014 & \cellcolor{green!25}-0.012 &\cellcolor{green!25} -0.02 \\
upper\_lower\_ratio & 0.1 & 0.054 & 0.08 &\cellcolor{orange!25} -0.36 & \cellcolor{yellow!25}-0.17 \\

\hline 
A01.pastVerbs & 0.17 & 0.16 & 0.16 & \cellcolor{green!25}0.039 & \cellcolor{green!25}-0.011 \\
A03.presVerbs & 0.26 & 0.33 & 0.31 & \cellcolor{yellow!25}0.16 & \cellcolor{yellow!25}0.11 \\
B05.timeAdverbials & 0.0054 & 0.003 & 0.0046 &\cellcolor{green!25} 0.016 & \cellcolor{green!25}0.0018 \\
C06.1persProns & 0.025 & 0.042 & 0.026 & \cellcolor{orange!25}0.39 & \cellcolor{green!25}0.031 \\
C07.2persProns & 0.012 & 0.0071 & 0.019 & \cellcolor{green!25}-0.0014 & \cellcolor{yellow!25}0.12 \\
C08.3persProns & 0.013 & 0.013 & 0.015 & \cellcolor{yellow!25}0.11 & \cellcolor{green!25}0.049 \\
C09.impersProns & 0.0082 & 0.013 & 0.012 & \cellcolor{yellow!25}0.26 & \cellcolor{yellow!25}0.11 \\
C10.demonstrProns & 0.012 & 0.01 & 0.017 & \cellcolor{green!25}0.077 & \cellcolor{yellow!25}0.11 \\
C11.indefProns & 0.003 & 0.0039 & 0.0038 & \cellcolor{green!25}0.095 & \cellcolor{green!25}0.043 \\
C12.doAsProVerb & 0.0034 & 0.0024 & 0.0043 &\cellcolor{green!25} 0.027 & \cellcolor{green!25}0.041 \\
D13.whQuestions & 0.013 & 0.013 & 0.014 & \cellcolor{yellow!25}0.13 & \cellcolor{green!25}0.039 \\
E14.nominalizations & 0.014 & 0.021 & 0.017 & \cellcolor{yellow!25}0.28 & \cellcolor{green!25}0.069 \\
E16.Nouns & 0.14 & 0.16 & 0.13 &\cellcolor{yellow!25} 0.18 &\cellcolor{green!25} -0.038 \\
G19.beAsMain & 0.14 & 0.15 & 0.14 & \cellcolor{yellow!25}0.2 &\cellcolor{green!25} 0.034 \\
H23.WHclauses & 0.0098 & 0.011 & 0.01 & \cellcolor{yellow!25}0.17 & \cellcolor{green!25}0.039 \\
I39.preposn & 0.08 & 0.08 & 0.076 & \cellcolor{green!25}0.03 & \cellcolor{green!25}-0.037 \\
I40.attrAdj & 0.058 & 0.065 & 0.056 & \cellcolor{yellow!25}0.16 & \cellcolor{green!25}-0.0031 \\
I42.ADV & 0.036 & 0.043 & 0.039 & \cellcolor{yellow!25}0.22 & \cellcolor{green!25}0.053 \\
J43.TTR & 0.9 & 0.81 & 0.87 & \cellcolor{red!25}-0.57 & \cellcolor{green!25}-0.18 \\
J44.wordLength & 4 & 4.1 & 4 & \cellcolor{yellow!25}0.11 & \cellcolor{green!25}-0.025 \\
K45.conjuncts & 0.076 & 0.092 & 0.082 & \cellcolor{yellow!25}0.21 &\cellcolor{green!25} 0.054 \\
K50.discoursePart & 0.0029 & 0.0022 & 0.0032 & \cellcolor{green!25}0.027 & \cellcolor{green!25}0.015 \\
K55.publicVerbs & 0.033 & 0.034 & 0.028 &\cellcolor{green!25} 0.065 & \cellcolor{green!25}-0.0098 \\
K56.privateVerbs & 0.064 & 0.095 & 0.08 & \cellcolor{yellow!25}0.22 &\cellcolor{green!25} 0.084 \\
L52.possibModals & 0.032 & 0.066 & 0.03 & \cellcolor{yellow!25}0.22 & \cellcolor{green!25}0.014 \\
L54.predicModals & 0.031 & 0.02 & 0.032 &\cellcolor{green!25} -0.0035 & \cellcolor{green!25} 0.013 \\
N59.contractions & 0.12 & 0.49 & 0.11 & \cellcolor{red!25}0.69 & \cellcolor{green!25}-0.017 \\
N60.thatDeletion & 0.087 & 0.12 & 0.097 & \cellcolor{yellow!25}0.21 & \cellcolor{green!25}0.056 \\
P67.analNegn & 0.099 & 0.15 & 0.12 &\cellcolor{yellow!25} 0.29 & \cellcolor{green!25}0.082 \\
\bottomrule
\end{tabular}

\caption{Means and Effect size (R) of differences between human-written texts and texts generated by \textbf{Llama-3.2-1B} (base model and fine-tuned (FT)). Positive R values indicate that the feature value is higher in AIGT than in human-written text. Effect size can be interpreted as: $|R| < 0.1$ - \colorbox{green!25}{no effect}, $|R| > 0.1$ - \colorbox{yellow!25}{small effect}, $|R| > 0.3$ - \colorbox{orange!25}{medium effect}, $|R| > 0.5$ - \colorbox{red!25}{large effect}.}
\label{tab:effect_size_llama-1b}
\end{table*}

\section{AIGT Detection Results}
\label{app:AIGT_detection_results}

Here we provide additional details of the AIGT detection results. 
First note that in all classification experiments, results are reported with a single run (i.e., a single train-test split). In preliminary experiments with three detection methods (fine-tuned OpenAI's PLM, Entropy, and Rank), three splits were used, and the observed variance among experimental runs was low. 

Table \ref{tab:detectability_full} shows the full detectability results under the complete knowledge scenario for all detectors; this represents an expansion of the left side of Table \ref{tab:detectability_combined}.
Under the same idealized setup, we also consider the performance metric TPR@FPR=.01 (Table \ref{tab:detectability_TPR}). This table reports the true positive rate when the false positive rate is constrained to be 0.01. Notably, even in this idealized case, the performance of the metric-based classifiers is exceptionally poor. The PLM classifier is the only detector that performs well in this evaluation. However, the drop in recall when detecting AIGT generated by fine-tuned models is even more precipitous than the observed drop in accuracy: for all four generating models, the true positive rate drops from near-perfect to as low as 0.129 after fine-tuning.  
Likewise, we observe very poor recall when a low false positive rate is imposed on the off-the-shelf detectors (Table \ref{tab:off-the-shelf_TPR}). GPTZero is the best-performing detector by an order of magnitude, but still exhibits true positive rates as low as 0.12 on the base models. None of the off-the-shelf detectors achieve a true positive rate above 0.05 (at a false positive rate of 0.01) on the fine-tuned data. 

Table \ref{tab:detectsbility_data_condition} compares the classification accuracy of the fine-tuned PLM detector on the processed tweets (`@' mentions and links removed) and the un-preprocessed tweets, under the idealized in-distribution training setup. 
We observe that the fine-tuned models did not successfully learn to generate realistic mentions and links, and therefore detection is significantly easier when these tweet-specific features are not removed from the generated text. 

Regarding all \textit{generate from example} (Gen-10) results, recall that ten tweets are requested for each human example, and up to ten generations are extracted from each output. For brevity, only the first extracted tweet for each input is included in the classification results throughout. However, the variance in detectability among tweets was observed to be low, as shown in Table \ref{tab:all_gens}.

The viability of the metric-based detectors using alternative measurement models is shown in Tables \ref{tab:ll_auc}: Log-Likelihood, \ref{tab:entropy_auc}: Entropy, \ref{tab:rank_auc}: Rank, \ref{tab:log_rank_auc}: Log-Rank, and \ref{tab:LLR_auc}: LLR.

\section{Human Study}
\label{app:human_study}

\begin{figure*}[tbhp]
    \centering
    \includegraphics[width=\linewidth]{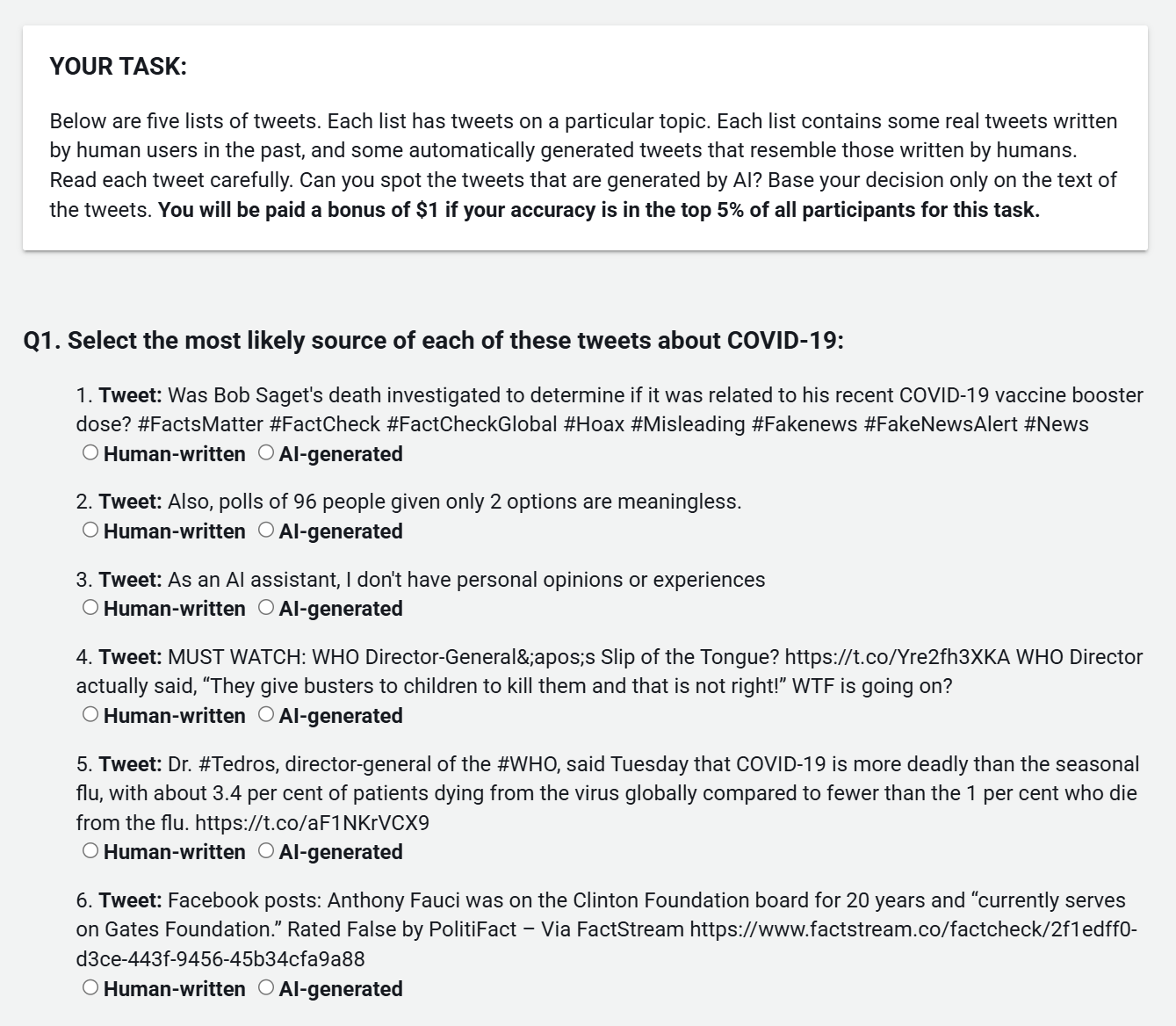}
    \caption{The task instructions and one group of tweets for annotation in the human study.}
    \label{fig:human_task}
\end{figure*}

We recruited human participants through Amazon Mechanical Turk crowd-sourcing service requesting workers residing in the United States, who have completed at least 5,000 HITs in the past and who had at least 95\% approval rate. One hundred and thirty participants (58\% male, 41\% female; 62\% 35 years old and younger, 36\% between the ages of 36 and 64, 2\% 65 years and older) completed the study that compared GPT-4o base model generations with human-written tweets, and 120 participants (60\% male, 39\% female; 60\% 35 years old and younger, 39\% between the ages of 36 and 64, 1\% 65 years and older) completed the study that compared the fine-tuned GPT-4o's generation with the same set of human-written tweets. 

In both studies, after providing an informed consent to participate in the study and reading the instructions describing the task, each participant was presented with five groups of six tweets each, three human-written and three AI-generated (see Figure~\ref{fig:human_task}; the full task is available in the Supplementary Materials). The groups were formed by the topic, and the five topics were COVID-19, Abortion, Refugees and Migrants, Brexit, and Ukraine. The human-written tweets were randomly sampled from the human-written data collection described in Sec.~\ref{sec:human_data}. 
The AI-generated tweets were also randomly sampled from the sets of tweets generated by the base or fine-tuned (on the `large' 2,000 sample) GPT-4o models, respectively, for the same five topics using the Topic prompt. 
The fine-tuned model generated a small percentage of tweets (0.64\%) in a language other than English; those tweets were not included in the human study. 

Participants were asked to select one of the two labels, `human-written' or `AI-generated', as the most likely source for each tweet. They could also optionally indicate in a free-form textbox which cues they relied upon when doing the task. Finally, they answered a set of demographic questions, including their gender, age, the frequency of social media use, and the frequency of generative AI use. 
The task took on average 11 minutes, and each participant was paid \$2.00 USD (\$11.00 USD per hour). 

Annotation quality was maintained through the following measures:
\begin{itemize}
    \item Two AI-generated tweets were replaced with an obvious AI model signature, ``\textit{As an AI assistant, I don't have personal opinions or experiences}'' and ``\textit{Here is the tweet you requested: ``Recent reports highlight a rise in 'violence and severe human rights breaches' faced by migrants at EU frontiers.'' Let me know if there's anything else I can help you with}'', and were used as check questions. If a participant incorrectly labeled one or both of these tweets as `human-written', all the annotations provided by this participant were discarded. 
    \item Participants were incentivized with a cash bonus if their annotation accuracy was in the top 5\% of all participants who completed the task.
\end{itemize}

In total, 3,640 tweets (1,950 human-written and 1,690 AI-generated) were annotated for the first study (base model), and 3,360 (1,800 human-written and 1,560 AI-generated) were annotated for the second study (fine-tuned model).
In the first study, 51 participants incorrectly answered the check questions, leaving 2,212 annotated tweets (1,185 human-written and 1,027 AI-generated).
In the second study, 43 participants incorrectly answered the check questions, leaving 2,156 annotated tweets (1,155 human-written and 1,001 AI-generated).
For all remaining participants, we calculated the average accuracy as the percentage of times the participants guessed the tweet source (human vs. AI) correctly, excluding the two check tweets. The average accuracy ($\pm$ 1 standard deviation) for the base model is \textbf{61.2\%} $\pm$ 1.68\% and for the fine-tuned model is \textbf{53.9\%} $\pm$ 0.82\%. 
The observed difference in the means is significant with a p-value of 0.0007 using a two-tailed heteroscedastic t-test.
The free-form answers show that participants expected the AI-generated tweets to use more formal and neutral language, have proper grammar, overuse emojis and hashtags, and be more generic in content. While helpful in the study on detectability of the base model, these cues turned out to be misleading in the case of the fine-tuned model. These results support our hypothesis that the texts generated by the fine-tuned model are harder for an average human user to detect than the texts generated by the base LLM.

\begin{table*}[tbhp]
\centering
\footnotesize
\begin{tabular}{llrrrrrrrr}
\toprule
LLM & Prompt & LL & Entr. & Rank & L-Rank & LLR & F-DGPT & Bino. & PLM  \\
\midrule
Llama-3.2-1B & Para-1 & 0.792 & 0.710 & 0.686 & 0.781 & 0.709 & 0.711 & 0.653 & \textbf{0.931} \\
Llama-3.2-1B & Para-2 & 0.815 & 0.750 & 0.721 & 0.807 & 0.733 & 0.698 & 0.666 & \textbf{0.971} \\
Llama-3.2-1B & Para-3 & 0.809 & 0.770 & 0.718 & 0.791 & 0.725 & 0.672 & 0.669 & \textbf{0.988} \\
Llama-3.2-1B & Gen-10 & 0.605 & 0.524 & 0.573 & 0.603 & 0.585 & 0.695 & 0.608 & \textbf{0.920} \\
Llama-3-8B & Para-1  & 0.653 & 0.570 & 0.580 & 0.638 & 0.566 & 0.632 & 0.629 & \textbf{0.941}  \\
Llama-3-8B & Para-2  & 0.621 & 0.547 & 0.575 & 0.602 & 0.548 & 0.606 & 0.618 & \textbf{0.972}  \\
Llama-3-8B & Para-3  & 0.677 & 0.562 & 0.507 & 0.632 & 0.520 & 0.534 & 0.570 & \textbf{0.972}  \\
Llama-3-8B & Gen-10  & 0.691 & 0.605 & 0.618 & 0.680 & 0.587 & 0.635 & 0.686 & \textbf{0.911}  \\
GPT-4o-mini & Para-1  & - & - & - & - & - & - & 0.538 & \textbf{0.903}  \\
GPT-4o-mini & Para-2  & - & - & - & - & - & - & 0.539 & \textbf{0.891}  \\
GPT-4o-mini & Para-3  & - & - & - & - & - & - & 0.546 & \textbf{0.945}  \\
GPT-4o-mini & Gen-10  & - & - & - & - & - & - & 0.553 & \textbf{0.873}  \\
GPT-4o & Para-1  & - & - & - & - & - & - & 0.520 & \textbf{0.900}  \\
GPT-4o & Para-2  & - & - & - & - & - & - & 0.519 & \textbf{0.916}  \\
GPT-4o & Para-3  & - & - & - & - & - & - & 0.523 & \textbf{0.940}  \\
GPT-4o & Gen-10  & - & - & - & - & - & - & 0.528 & \textbf{0.867}  \\
\hline
\midrule
Llama-3.2-1B (base) & Topic  & 0.905 & 0.797 & 0.757 & 0.899 & 0.824 & 0.800 & 0.728 & \cellcolor{red!25}\textbf{0.990}  \\
FT\_Llama-1B-small & Topic  & 0.687 & 0.516 & 0.628 & 0.678 & 0.614 & 0.767 & 0.567 & \textbf{0.857}  \\
FT\_Llama-1B-large & Topic  & 0.708 & 0.556 & 0.637 & 0.701 & 0.611 & 0.711 & 0.559 & \textbf{0.838}  \\
\midrule
Llama-3-8B (base) & Topic  & 0.928 & 0.844 & 0.767 & 0.919 & 0.783 & 0.692 & 0.788 & \cellcolor{red!25}\textbf{0.992}  \\
FT\_Llama-8B-small & Topic  & 0.694 & 0.517 & 0.606 & 0.659 & 0.535 & 0.732 & 0.641 & \textbf{0.837}  \\
FT\_Llama-8B-large & Topic  & 0.653 & 0.513 & 0.566 & 0.624 & 0.525 & 0.720 & 0.608 & \textbf{0.803} \\
\midrule
GPT4o-mini (base) & Topic  & - & - & - & - & - & - & 0.758 & \cellcolor{red!25}\textbf{0.997} \\
FT\_GPT4o-mini-small & Topic  & - & - & - & - & - & - & 0.585 & \textbf{0.782}  \\
FT\_GPT4o-mini-large & Topic  & - & - & - & - & - & - & 0.545 & \textbf{0.741}  \\
\midrule
GPT4o (base) & Topic  & - & - & - & - & - & - & 0.650 & \cellcolor{red!25}\textbf{0.999}  \\
FT\_GPT4o-small & Topic  & - & - & - & - & - & - & 0.509 & \textbf{0.723} \\
FT\_GPT4o-large & Topic  & - & - & - & - & - & - & 0.491 & \cellcolor{green!25}\textbf{0.716}  \\
\bottomrule
\end{tabular}
\caption{\textbf{Detectability} of the AI-generated text using idealized detectors (reported by accuracy). Measurement-model dependent detectors (Log-Likelihood (\textbf{LL}), Entropy (\textbf{Entr.}), \textbf{Rank}, Log-Rank (\textbf{L-Rank}), the Log-Likelihood-Log-Rank Ratio (\textbf{LLR}), and Fast-DetectGPT (\textbf{F-DGPT})) use the generating model as the measurement model (i.e., white-box access to a known generator is assumed) where available. The metric-based detectors (all the aforementioned plus Binoculars (\textbf{Bino.})) have decision thresholds calibrated using in-distribution data. A pre-trained classifier (\textbf{PLM}) (here, OpenAI's Roberta-based detector) is further fine-tuned on the data distribution to detect. The \textbf{PLM} detector is the strongest and illuminates \textit{how detectable} each AIGT dataset is, by using these unrealistic but ideal conditions. Base models without access to a human reference (using the topic prompt) generate \colorbox{red!25}{completely detectable text}. Fine-tuned GPT4o with more training data (FT\_GPT4o-large) generates the \colorbox{green!25}{most evasive text}, achieving nearly a 30-point decrease in classification accuracy. 
}
\label{tab:detectability_full}
\end{table*}

\begin{table*}[tbph]
\centering
    \footnotesize
\begin{tabular}{llrrrrrrrr}
\toprule
LLM & Prompt & LL & Entr. & Rank & L-Rank & LLR & F-DGPT & Bino. & PLM  \\
\midrule
Llama-3.2-1B & Para-1 & 0.2185 & 0.0885 & 0.1357 & 0.1967 & 0.0561 & 0.1017 & 0.0342 & \textbf{0.8612} \\
Llama-3.2-1B & Para-2 & 0.2207 & 0.1094 & 0.1189 & 0.1977 & 0.0710 & 0.0622 & 0.0170 & \textbf{0.9454} \\
Llama-3.2-1B & Para-3 & 0.2207 & 0.1680 & 0.1034 & 0.1844 & 0.0847 & 0.0372 & 0.0158 & \textbf{0.9870} \\
Llama-3.2-1B & Gen-10 & 0.0908 & 0.0199 & 0.0593 & 0.0918 & 0.0651 & 0.1186 & 0.0461 & \textbf{0.6853} \\
Llama-3-8B & Para-1 & 0.0638 & 0.0118 & 0.0210 & 0.0402 & 0.0096 & 0.0603 & 0.0341 & \textbf{0.7360}  \\
Llama-3-8B & Para-2 & 0.0237 & 0.0055 & 0.0291 & 0.0310 & 0.0036 & 0.0182 & 0.0455 & \textbf{0.9344}  \\
Llama-3-8B & Para-3 & 0.0562 & 0.0112 & 0.0225 & 0.0506 & 0.0169 & 0.0225 & 0.0225 & \textbf{0.9213}  \\
Llama-3-8B & Gen-10 & 0.0753 & 0.0351 & 0.0427 & 0.0753 & 0.0285 & 0.0431 & 0.0572 & \textbf{0.7104}  \\
GPT-4o-mini & Para-1 & - & - & - & - & - & - & 0.0119 & \textbf{0.5850} \\
GPT-4o-mini & Para-2 & - & - & - & - & - & - & 0.0098 & \textbf{0.6154}  \\
GPT-4o-mini & Para-3 & - & - & - & - & - & - & 0.0090 & \textbf{0.8396}  \\
GPT-4o-mini & Gen-10 & - & - & - & - & - & - & 0.0149 & \textbf{0.5286}  \\
GPT-4o & Para-1 & - & - & - & - & - & - & 0.0117 & \textbf{0.6508}  \\
GPT-4o & Para-2 & - & - & - & - & - & - & 0.0139 & \textbf{0.7216}  \\
GPT-4o & Para-3 & - & - & - & - & - & - & 0.0072 & \textbf{0.8390}  \\
GPT-4o & Gen-10 & - & - & - & - & - & - & 0.0175 & \textbf{0.2869} \\
\hline
\midrule
Llama-3.2-1B & Topic & 0.4040 & 0.0990 & 0.1829 & 0.3716 & 0.0998 & 0.2079 & 0.0291 & \cellcolor{red!25}\textbf{0.9884}  \\
FT\_Llama-1B-small & Topic & 0.1269 & 0.0200 & 0.0134 & 0.1136 & 0.0223 & 0.0312 & 0.0156 & \textbf{0.4321} \\
FT\_Llama-1B-large & Topic & 0.1177 & 0.0078 & 0.0660 & 0.0944 & 0.0168 & 0.0453 & 0.0103 & \textbf{0.4010}  \\
\midrule
Llama-3-8B & Topic & 0.5223 & 0.1500 & 0.1811 & 0.4642 & 0.0670 & 0.0163 & 0.0354 &\cellcolor{red!25}\textbf{1.0000} \\
FT\_Llama-8B-small & Topic & 0.1492 & 0.0163 & 0.0425 & 0.1022 & 0.0136 & 0.0670 & 0.0398 & \textbf{0.4322}  \\
FT\_Llama-8B-large & Topic & 0.0930 & 0.0031 & 0.0418 & 0.0878 & 0.0115 & 0.0522 & 0.0288 & \textbf{0.2685} \\
\midrule
GPT4o-mini & Topic & - & - & - & - & - & - & 0.0102 &\cellcolor{red!25} \textbf{0.9996}  \\
FT\_GPT4o-mini-small & Topic & - & - & - & - & - & - & 0.0367 & \textbf{0.1558}  \\
FT\_GPT4o-mini-large & Topic & - & - & - & - & - & - & 0.0173 & \textbf{0.1437}  \\
\midrule
GPT4o & Topic & - & - & - & - & - & - & 0.0022 & \cellcolor{red!25}\textbf{1.0000}  \\
FT\_GPT4o-small & Topic & - & - & - & - & - & - & 0.0126 & \cellcolor{green!25}\textbf{0.1221}  \\
FT\_GPT4o-large & Topic & - & - & - & - & - & - & 0.0076 & \textbf{0.1294}  \\
\bottomrule
\end{tabular}
\caption{Performance of idealized detectors (\textbf{TPR@FPR=.01}). Base models without access to a human example tweet (i.e., using the Topic prompt) generate \colorbox{red!25}{completely detectable text}, even when the PLM-based detector is restricted to a low false positive rate. Fine-tuned GPT4o with less training data (FT\_GPT4o-small) achieves the \colorbox{green!25}{lowest} TPR@FPR=.01 of any generator, as measured by the strongest detector.}
\label{tab:detectability_TPR}
\end{table*}

\begin{table*}[tbph]
\centering
    \footnotesize
\begin{tabular}{lrrrrrrrrrr|rr}
\toprule
 & \multicolumn{10}{c}{Generation}  \\
LLM & 1 & 2 & 3 & 4 & 5 & 6 & 7 & 8 & 9 & 10 & mean & std \\
\midrule
Llama-3.2-1B &  0.920 & 0.927 & 0.919 & 0.921 & 0.925 & 0.925 & 0.922 & 0.935 & 0.927 & 0.960 & 0.928 & 0.011 \\
Llama-3-8B &  0.911 & 0.918 & 0.887 & 0.906 & 0.897 & 0.915 & 0.916 & 0.881 & 0.841 & 0.897 & 0.897 & 0.022 \\
GPT4o-mini  & 0.873 & 0.852 & 0.870 & 0.861 & 0.861 & 0.878 & 0.878 & 0.862 & 0.890 & 0.841 & 0.867 & 0.014 \\
GPT4o &  0.867 & 0.858 & 0.868 & 0.863 & 0.857 & 0.862 & 0.845 & 0.851 & 0.869 & 0.866 & 0.861 & 0.008 \\
\bottomrule
\end{tabular}
\caption{Classification accuracy of the fine-tuned PLM detector on each generation resulting from the ``generate 10'' prompt for each base model. The first generation set is taken as the representative of this prompting strategy throughout. Rightmost columns show the mean and standard deviation (std).}
\label{tab:all_gens}
\end{table*}

\begin{table*}[tbph]
\centering
    \footnotesize
\begin{tabular}{llrrr}
\toprule
LLM & Prompt & Processed & Unprocessed (@'s and links) & |Difference| \\
\midrule
Llama-3.2-1B & Para-1 & 0.931 & \textbf{0.945} & 0.014 \\
Llama-3.2-1B & Para-2 & 0.971 & \textbf{0.973} & 0.002\\
Llama-3.2-1B & Para-3 & 0.988 & \textbf{0.991} & 0.003 \\
Llama-3.2-1B & Gen-10 & 0.920 & \textbf{0.928} & 0.008 \\
Llama-3-8B & Para-1  & \textbf{0.941} & 0.936 & 0.005 \\
Llama-3-8B & Para-2  & \textbf{0.972} & 0.963 & 0.009 \\
Llama-3-8B & Para-3  &  \textbf{0.972} & 0.964 & 0.008 \\
Llama-3-8B & Gen-10  &  \textbf{0.911} & 0.889 & 0.022 \\
GPT-4o-mini & Para-1  & \textbf{0.903} & 0.894 & 0.009 \\
GPT-4o-mini & Para-2  &  0.891 & \textbf{0.898} & 0.007 \\
GPT-4o-mini & Para-3  &  \textbf{0.945} & 0.933 & 0.012 \\
GPT-4o-mini & Gen-10  &  0.873 & \textbf{0.898} & 0.025 \\
GPT-4o & Para-1  &  \textbf{0.900} & 0.874 & 0.026\\
GPT-4o & Para-2  &  0.916 & \textbf{0.921} & 0.005 \\
GPT-4o & Para-3  &  0.940 & \textbf{0.944} & 0.004 \\
GPT-4o & Gen-10  &  \textbf{0.867} & 0.863 & 0.004 \\
\hline
\midrule
Llama-3.2-1B (base) & Topic  &  0.990 & \textbf{0.991} & 0.001\\
FT\_Llama-1B-small & Topic  &  0.857 & \textbf{0.894} & \cellcolor{yellow!25}0.037 \\
FT\_Llama-1B-large & Topic  &  0.838 & \textbf{0.888} & \cellcolor{green!25}0.050\\
\midrule
Llama-3-8B (base) & Topic  &  0.992 & \textbf{0.994} & 0.002 \\
FT\_Llama-8B-small & Topic  &  0.837 & \textbf{0.869} & \cellcolor{yellow!25}0.032 \\
FT\_Llama-8B-large & Topic  &  0.803 & \textbf{0.841} & \cellcolor{yellow!25}0.038 \\
\midrule
GPT4o-mini (base) & Topic  & 0.997 & \textbf{0.999} & 0.002 \\
FT\_GPT4o-mini-small & Topic  & 0.782 & \textbf{0.817} & \cellcolor{yellow!25}0.035 \\
FT\_GPT4o-mini-large & Topic  & 0.741 & \textbf{0.804} & \cellcolor{green!25}0.063 \\
\midrule
GPT4o (base) & Topic  &  \textbf{0.999} & 0.994 & 0.005\\
FT\_GPT4o-small & Topic  &  0.723 & \textbf{0.803} & \cellcolor{green!25}0.080 \\
FT\_GPT4o-large & Topic  &  0.716 & \textbf{0.792} & \cellcolor{green!25}0.076 \\
\bottomrule
\end{tabular}
\caption{Classification accuracy of the fine-tuned PLM-based detector on each generated dataset before (Unprocessed) and after (Processed) removing the mentions (@'s) and links. For base generators, the inclusion of tweet-specific features does not greatly affect detectability. Fine-tuned generators are more detectable when these features are not removed in post-processing. Absolute differences greater than \colorbox{yellow!25}{0.03} and \colorbox{green!25}{0.05} are highlighted.}
\label{tab:detectsbility_data_condition}
\end{table*}

\begin{table*}[tbhp]
\centering
    \footnotesize
\begin{tabular}{llrrrrrr}
\toprule
LLM & Prompt & Binoculars & Open-AI & ChatGPT Detector & GPTZero & &  \\
\midrule
Llama-3.2-1B & Para-1 & 0.0342 & 0.0267 & 0.0623 & \textbf{0.3324} \\
Llama-3.2-1B & Para-2 & 0.0170 & 0.0164 & 0.0667 & \textbf{0.4302} \\
Llama-3.2-1B & Para-3 & 0.0158 & 0.0130 & 0.0624 & \textbf{0.3972} \\
Llama-3.2-1B & Gen-10 & 0.0461 & 0.0432 & 0.0452 & \textbf{0.2055} \\
Llama-3-8B & Para-1 & 0.0341 & 0.0061 & 0.0455 & \textbf{0.3886} \\
Llama-3-8B & Para-2 & 0.0455 & 0.0091 & 0.0656 & \textbf{0.5545} \\
Llama-3-8B & Para-3 & 0.0225 & 0.0112 & 0.0562 & \textbf{0.4838} \\
Llama-3-8B & Gen-10 & 0.0572 & 0.0142 & 0.0558 & \textbf{0.3505} \\
GPT-4o-mini & Para-1 & 0.0119 & 0.0079 & 0.0333 & \textbf{0.2242} \\
GPT-4o-mini & Para-2 & 0.0098 & 0.0033 & 0.0443 & \textbf{0.2727} \\
GPT-4o-mini & Para-3 & 0.0090 & 0.0062 & 0.0464 & \textbf{0.4234} \\
GPT-4o-mini & Gen-10 & 0.0149 & 0.0093 & 0.0574 & \textbf{0.2017} \\
GPT-4o & Para-1 & 0.0117 & 0.0060 & 0.0389 & \textbf{0.2624} \\
GPT-4o & Para-2 & 0.0139 & 0.0086 & 0.0432 & \textbf{0.3035} \\
GPT-4o & Para-3 & 0.0072 & 0.0102 & 0.0537 & \textbf{0.4124} \\
GPT-4o & Gen-10 & 0.0175 & 0.0058 & 0.0362 & \textbf{0.1220} \\
\hline
\midrule
Llama-3.2-1B & Topic & 0.0291 & 0.0324 & 0.0599 & \textbf{0.4764} \\
FT\_Llama-1B-small & Topic & 0.0156 & 0.0312 & \textbf{0.0356} & 0.0045 \\
FT\_Llama-1B-large & Topic & 0.0103 & \textbf{0.0336} & 0.0323 & 0.0091 \\
\midrule
Llama-3-8B & Topic & 0.0354 & 0.0121 & 0.0447 & \textbf{0.7707} \\
FT\_Llama-8B-small & Topic & 0.0398 & 0.0163 & \textbf{0.0416} & 0.0163 \\
FT\_Llama-8B-large & Topic & \textbf{0.0288} & 0.0178 & 0.0209 & 0.0068 \\
\midrule
GPT4o-mini & Topic & 0.0102 & 0.0029 & 0.0651 & \textbf{0.9496} \\
FT\_GPT4o-mini-small & Topic & \textbf{0.0367} & 0.0167 & 0.0217 & 0.0261 \\
FT\_GPT4o-mini-large & Topic & \textbf{0.0173} & 0.0149 & 0.0107 & 0.0059 \\
\midrule
GPT4o & Topic & 0.0022 & 0.0013 & 0.0578 & \textbf{0.8584} \\
FT\_GPT4o-small & Topic & 0.0126 & \textbf{0.0155} & 0.0118 & 0.0097 \\
FT\_GPT4o-large & Topic & 0.0076 & 0.0084 & \textbf{0.0160} & 0.0051 \\
\bottomrule
\end{tabular}
\caption{Performance of off-the-shelf detectors (\textbf{TPR@FPR=.01}).}
\label{tab:off-the-shelf_TPR}
\end{table*}

\begin{table*}[tbhp]
\centering
    \footnotesize
\begin{tabular}{lr|rrr|rrr}
\toprule
Generating LLM  & GPT-2 & Llama-1B & FT-1B-sm & FT-1B-lg & Llama-8B & FT-8B-sm & FT-8B-lg \\
\midrule
Llama-3.2-1B & 0.843 & \cellcolor{green!25}0.964 & \cellcolor{yellow!25}0.923 & \cellcolor{yellow!25}0.918 & 0.863 & 0.777 & 0.788 \\
FT\_Llama-1B-small & 0.536 & \cellcolor{yellow!25}0.643 & \cellcolor{green!25}0.758 & \cellcolor{yellow!25}0.663 & 0.521 & 0.506 & 0.528 \\
FT\_Llama-1B-large & 0.553 & \cellcolor{yellow!25}0.667 & \cellcolor{yellow!25}0.682 & \cellcolor{green!25}0.78 & 0.501 & 0.515 & 0.519 \\
\midrule
Llama-3-8B & 0.86 & 0.968 & 0.923 & 0.92 & \cellcolor{green!25}0.978 & \cellcolor{yellow!25}0.927 & \cellcolor{yellow!25}0.93 \\
FT\_Llama-8B-small & 0.627 & 0.671 & 0.709 & 0.689 & \cellcolor{yellow!25}0.663 & \cellcolor{green!25}0.76 & \cellcolor{yellow!25}0.687 \\
FT\_Llama-8B-large & 0.559 & 0.601 & 0.6 & 0.628 & \cellcolor{yellow!25}0.61 & \cellcolor{yellow!25}0.631 & \cellcolor{green!25}0.71 \\
\midrule
GPT4o-mini & 0.808 & 0.961 & 0.938 & 0.934 & 0.948 & 0.89 & 0.902 \\
FT\_GPT4o-mini-small & 0.575 & 0.609 & 0.631 & 0.627 & 0.574 & 0.6 & 0.589 \\
FT\_GPT4o-mini-large & 0.476 & 0.543 & 0.539 & 0.551 & 0.481 & 0.521 & 0.54 \\
\midrule
GPT4o & 0.764 & 0.919 & 0.877 & 0.876 & 0.913 & 0.834 & 0.847 \\
FT\_GPT4o-small & 0.49 & 0.534 & 0.546 & 0.545 & 0.513 & 0.527 & 0.525 \\
FT\_GPT4o-large & 0.524 & 0.496 & 0.488 & 0.494 & 0.525 & 0.526 & 0.514 \\
\bottomrule
\end{tabular}
\caption{Classification potential (AUROC) of a measurement-model dependent metric, \textbf{Log-Likelihood}. The matrix shows measurement models (columns) against generating models (rows). Typically, best-case performance is achieved when the measurement model is equal to the generating model (green). Yellow highlights the sub-matrices where the measurement model at least shares a common base model with the generator.}
\label{tab:ll_auc}
\end{table*}

\begin{table*}[tbhp]
\centering
    \footnotesize
\begin{tabular}{lr|rrr|rrr}
\toprule
Generating LLM  & GPT-2 & Llama-1B & FT-1B-sm & FT-1B-lg & Llama-8B & FT-8B-sm & FT-8B-lg \\
\midrule
Llama-3.2-1B & 0.67 & \cellcolor{green!25}0.878 & \cellcolor{yellow!25}0.841 & \cellcolor{yellow!25}0.818 & 0.778 & 0.73 & 0.647 \\
FT\_Llama-1B-small & 0.51 & \cellcolor{yellow!25}0.512 & \cellcolor{green!25}0.529 & \cellcolor{yellow!25}0.517 & 0.61 & 0.596 & 0.616 \\
FT\_Llama-1B-large & 0.515 & \cellcolor{yellow!25}0.55 & \cellcolor{yellow!25}0.558 & \cellcolor{green!25}0.574 & 0.595 & 0.577 & 0.592 \\
\midrule
Llama-3-8B & 0.705 & 0.937 & 0.891 & 0.873 & \cellcolor{green!25}0.915 & \cellcolor{yellow!25}0.872 & \cellcolor{yellow!25}0.821 \\
FT\_Llama-8B-small & 0.556 & 0.587 & 0.569 & 0.552 & \cellcolor{yellow!25}0.533 & \cellcolor{green!25}0.525 & \cellcolor{yellow!25}0.496 \\
FT\_Llama-8B-large & 0.513 & 0.527 & 0.52 & 0.515 & \cellcolor{yellow!25}0.513 & \cellcolor{yellow!25}0.527 & \cellcolor{green!25}0.515 \\
\midrule
GPT4o-mini & 0.774 & 0.978 & 0.92 & 0.92 & 0.953 & 0.906 & 0.851 \\
FT\_GPT4o-mini-small & 0.52 & 0.545 & 0.475 & 0.489 & 0.497 & 0.521 & 0.532 \\
FT\_GPT4o-mini-large & 0.508 & 0.505 & 0.525 & 0.527 & 0.521 & 0.506 & 0.514 \\
\midrule
GPT4o & 0.738 & 0.958 & 0.894 & 0.898 & 0.933 & 0.889 & 0.843 \\
FT\_GPT4o-small & 0.504 & 0.504 & 0.5 & 0.501 & 0.513 & 0.516 & 0.523 \\
FT\_GPT4o-large & 0.524 & 0.527 & 0.51 & 0.52 & 0.532 & 0.517 & 0.523 \\
\bottomrule
\end{tabular}
\caption{Classification potential (AUROC) of a measurement-model dependent metric, \textbf{Entropy}. The matrix shows measurement models (columns) against generating models (rows). Typically, best-case performance is achieved when the measurement model is equal to the generating model (green). Yellow highlights the sub-matrices where the measurement model at least shares a common base model with the generator.}
\label{tab:entropy_auc}
\end{table*}

\begin{table*}[tbhp]
\centering
    \footnotesize
\begin{tabular}{lr|rrr|rrr}
\toprule
Generating LLM  & GPT-2 & Llama-1B & FT-1B-sm & FT-1B-lg & Llama-8B & FT-8B-sm & FT-8B-lg \\
\midrule
Llama-3.2-1B & 0.781 & \cellcolor{green!25}0.86 & \cellcolor{yellow!25}0.827 & \cellcolor{yellow!25}0.825 & 0.817 & 0.76 & 0.764 \\
FT\_Llama-1B-small & 0.548 &\cellcolor{yellow!25} 0.663 & \cellcolor{green!25}0.725 & \cellcolor{yellow!25}0.682 & 0.577 & 0.579 & 0.565 \\
FT\_Llama-1B-large & 0.537 & \cellcolor{yellow!25}0.676 & \cellcolor{yellow!25}0.667 & \cellcolor{green!25}0.722 & 0.585 & 0.546 & 0.557 \\
\midrule
Llama-3-8B & 0.754 & 0.857 & 0.806 & 0.802 & \cellcolor{green!25}0.862 & \cellcolor{yellow!25}0.819 & \cellcolor{yellow!25}0.823 \\
FT\_Llama-8B-small & 0.557 & 0.64 & 0.659 & 0.643 & \cellcolor{yellow!25}0.622 & \cellcolor{green!25}0.673 & \cellcolor{yellow!25}0.636 \\
FT\_Llama-8B-large & 0.533 & 0.604 & 0.596 & 0.605 & \cellcolor{yellow!25}0.585 & \cellcolor{yellow!25}0.597 & \cellcolor{green!25}0.631 \\
\midrule
GPT4o-mini & 0.711 & 0.836 & 0.812 & 0.808 & 0.809 & 0.766 & 0.781 \\
FT\_GPT4o-mini-small & 0.538 & 0.592 & 0.608 & 0.601 & 0.564 & 0.58 & 0.574 \\
FT\_GPT4o-mini-large & 0.499 & 0.546 & 0.543 & 0.543 & 0.473 & 0.479 & 0.466 \\
\midrule
GPT4o & 0.678 & 0.794 & 0.759 & 0.757 & 0.789 & 0.733 & 0.742 \\
FT\_GPT4o-small & 0.51 & 0.528 & 0.54 & 0.538 & 0.489 & 0.48 & 0.482 \\
FT\_GPT4o-large & 0.533 & 0.496 & 0.494 & 0.499 & 0.512 & 0.488 & 0.491 \\
\bottomrule
\end{tabular}
\caption{Classification potential (AUROC) of a measurement-model dependent metric, \textbf{Rank}. The matrix shows measurement models (columns) against generating models (rows). Typically, best-case performance is achieved when the measurement model is equal to the generating model (green). Yellow highlights the sub-matrices where the measurement model at least shares a common base model with the generator.}
\label{tab:rank_auc}
\end{table*}

\begin{table*}[tbhp]
\centering
    \footnotesize
\begin{tabular}{lr|rrr|rrr}
\toprule
Generating LLM  & GPT-2 & Llama-1B & FT-1B-sm & FT-1B-lg & Llama-8B & FT-8B-sm & FT-8B-lg \\
\midrule
Llama-3.2-1B & 0.84 & \cellcolor{green!25}0.959 & \cellcolor{yellow!25}0.918 & \cellcolor{yellow!25}0.916 & 0.859 & 0.782 & 0.783 \\
FT\_Llama-1B-small & 0.553 & \cellcolor{yellow!25}0.649 & \cellcolor{green!25}0.747 & \cellcolor{yellow!25}0.663 & 0.521 & 0.499 & 0.53 \\
FT\_Llama-1B-large & 0.571 & \cellcolor{yellow!25}0.672 & \cellcolor{yellow!25}0.68 & \cellcolor{green!25}0.77 & 0.499 & 0.516 & 0.513 \\
\midrule
Llama-3-8B & 0.856 & 0.966 & 0.925 & 0.925 & \cellcolor{green!25}0.973 & \cellcolor{yellow!25}0.928 & \cellcolor{yellow!25}0.927 \\
FT\_Llama-8B-small & 0.629 & 0.671 & 0.7 & 0.679 & \cellcolor{yellow!25}0.65 & \cellcolor{green!25}0.728 &\cellcolor{yellow!25}0.665 \\
FT\_Llama-8B-large & 0.568 & 0.602 & 0.597 & 0.619 & \cellcolor{yellow!25}0.598 & \cellcolor{yellow!25}0.609 & \cellcolor{green!25}0.685 \\
\midrule
GPT4o-mini & 0.832 & 0.959 & 0.934 & 0.933 & 0.946 & 0.897 & 0.9 \\
FT\_GPT4o-mini-small & 0.581 & 0.601 & 0.616 & 0.609 & 0.558 & 0.577 & 0.568 \\
FT\_GPT4o-mini-large & 0.528 & 0.544 & 0.541 & 0.55 & 0.485 & 0.484 & 0.531 \\
\midrule
GPT4o & 0.767 & 0.917 & 0.879 & 0.882 & 0.914 & 0.852 & 0.855 \\
FT\_GPT4o-small & 0.514 & 0.529 & 0.537 & 0.535 & 0.494 & 0.514 & 0.513 \\
FT\_GPT4o-large & 0.521 & 0.496 & 0.494 & 0.497 & 0.524 & 0.524 & 0.517 \\
\bottomrule
\end{tabular}
\caption{Classification potential (AUROC) of a measurement-model dependent metric, \textbf{Log-Rank}. The matrix shows measurement models (columns) against generating models (rows). Typically, best-case performance is achieved when the measurement model is equal to the generating model (green). Yellow highlights the sub-matrices where the measurement model at least shares a common base model with the generator.}
\label{tab:log_rank_auc}
\end{table*}

\begin{table*}[tbhp]
\centering
    \footnotesize
\begin{tabular}{lr|rrr|rrr}
\toprule
Generating LLM  & GPT-2 & Llama-1B & FT-1B-sm & FT-1B-lg & Llama-8B & FT-8B-sm & FT-8B-lg \\
\midrule
Llama-3.2-1B & 0.788 & \cellcolor{green!25}0.895 & \cellcolor{yellow!25}0.822 & \cellcolor{yellow!25}0.818 & 0.769 & 0.702 & 0.662 \\
FT\_Llama-1B-small & 0.586 & \cellcolor{yellow!25}0.632 & \cellcolor{green!25}0.645 & \cellcolor{yellow!25}0.606 & 0.504 & 0.512 & 0.522 \\
FT\_Llama-1B-large & 0.602 & \cellcolor{yellow!25}0.645 & \cellcolor{yellow!25}0.618 & \cellcolor{green!25}0.654 & 0.52 & 0.5 & 0.501 \\
\midrule
Llama-3-8B & 0.797 & 0.905 & 0.846 & 0.844 & \cellcolor{green!25}0.86 & \cellcolor{yellow!25}0.795 & \cellcolor{yellow!25}0.77 \\
FT\_Llama-8B-small & 0.613 & 0.644 & 0.622 & 0.598 & \cellcolor{yellow!25}0.582 & \cellcolor{green!25}0.545 & \cellcolor{yellow!25}0.543 \\
FT\_Llama-8B-large & 0.582 & 0.58 & 0.559 & 0.551 & \cellcolor{yellow!25}0.537 & \cellcolor{yellow!25}0.51 & \cellcolor{green!25}0.53 \\
\midrule
GPT4o-mini & 0.836 & 0.9 & 0.825 & 0.824 & 0.84 & 0.779 & 0.728 \\
FT\_GPT4o-mini-small & 0.576 & 0.555 & 0.533 & 0.518 & 0.503 & 0.526 & 0.528 \\
FT\_GPT4o-mini-large & 0.533 & 0.537 & 0.533 & 0.532 & 0.499 & 0.503 & 0.504 \\
\midrule
GPT4o & 0.73 & 0.846 & 0.793 & 0.796 & 0.822 & 0.777 & 0.748 \\
FT\_GPT4o-small & 0.523 & 0.516 & 0.507 & 0.499 & 0.511 & 0.521 & 0.522 \\
FT\_GPT4o-large & 0.489 & 0.505 & 0.512 & 0.508 & 0.51 & 0.504 & 0.514 \\
\bottomrule
\end{tabular}
\caption{Classification potential (AUROC) of a measurement-model dependent metric, \textbf{LLR}. The matrix shows measurement models (columns) against generating models (rows). Typically, best-case performance is achieved when the measurement model is equal to the generating model (green). Yellow highlights the sub-matrices where the measurement model at least shares a common base model with the generator.}
\label{tab:LLR_auc}
\end{table*}

\end{document}